\pdfoutput=1
\documentclass[11pt]{article}

\usepackage[preprint]{acl}
\usepackage{times}
\usepackage{latexsym}
\usepackage[T1]{fontenc}
\usepackage[utf8]{inputenc}
\usepackage[expansion=false]{microtype}
\usepackage{inconsolata}
\usepackage{graphicx}
\usepackage{xurl}%
\usepackage{amsmath}
\usepackage{amssymb}
\usepackage{booktabs}
\usepackage{array}
\usepackage{multirow}
\usepackage{xcolor}
\usepackage{colortbl}%
\usepackage{pifont}%
\definecolor{arborabg}{HTML}{E7F2F8}\definecolor{arborablue}{HTML}{0072B2}\definecolor{cmgreen}{HTML}{00866A}\definecolor{xmgrey}{HTML}{9AA3AD}

\usepackage{capt-of}
\usepackage{enumitem}%
\usepackage{algorithm}
\usepackage[noend]{algpseudocode}

\newif\ifwithappendix
\newif\ifpreprintlayout
\newcommand{\headgap}{\hspace{0pt plus 0.3em}}
\newcommand{\looseacl}{\ifpreprintlayout\else\looseness=1 \fi}
\newcommand{\lineadj}[2]{\ifpreprintlayout\looseness=#2 \else\looseness=#1 \fi}
\newcommand{\appref}[1]{\ifwithappendix#1\fi}

\newif\ifpendshow \pendshowtrue

\newcommand{\arbora}{Arbora}
\newcommand{\aidisclosure}{We used AI assistants for coding and for editing the text. The research ideas and all claims are the authors' own, and all results come from real experiments.}
\newcommand{\arboraE}{\arbora{}}
\newcommand{\arboraOne}{\arbora{} w/o experts}
\newcommand{\arboraEp}{\arbora{} w/o path scorer}
\newcommand{\gainALgreedy}{25.5\%}
\newcommand{\gainTPgreedy}{21.6\%}
\newcommand{\gainALsample}{27.6\%}
\newcommand{\gainTPsample}{23.9\%}

\newcommand{\overviewfigure}{%
  \includegraphics[width=\textwidth]{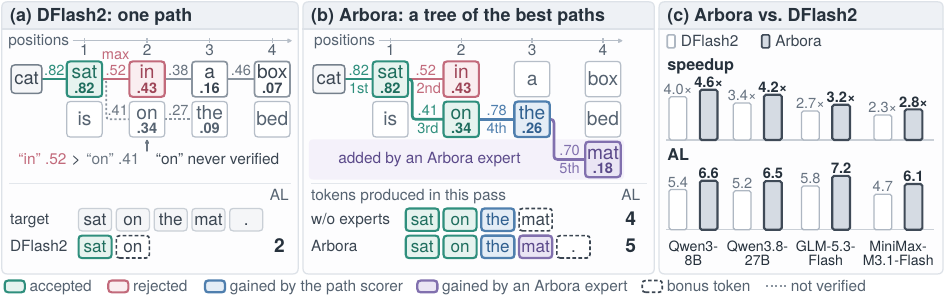}\par
  \captionsetup{hypcap=false}%
  \captionof{figure}{(a)~DFlash2 keeps only the best next token at each position, so the target verifies one path and rejects ``in'', although ``on'' is a candidate. (b)~\arbora{}'s path scorer ranks whole paths, and the target verifies a tree of the five best paths in one pass. The path ``sat on'' ranks 3rd, so the target also verifies ``on'' and accepts it. An Arbora expert adds ``mat''. A number in a cell is the probability of the path up to that token, and a number on an edge is the probability of that token after the previous one, from DFlash2 in (a) and the path scorer in (b). These probabilities and the number of verified paths are illustrative. (c)~Speedup over the target alone and accept length (AL) at temperature $T=0$, averaged over the six rows of Table~\ref{tab:main}.}
  \label{fig:overview}\par
  \vskip\dbltextfloatsep}
\newcommand{\maketitlewithoverview}{%
  \let\originaltwocolumn\twocolumn
  \renewcommand{\twocolumn}[1][]{\originaltwocolumn[##1\overviewfigure]}%
  \maketitle
  \let\twocolumn\originaltwocolumn}

\newcommand{\papertitle}{Arbora: Candidate-Aware Mixture of Speculators\\ for Block Diffusion Drafters}
\title{\papertitle}

\withappendixtrue
\preprintlayouttrue

\newcommand{\codelink}{\href{https://github.com/FutureMLS-Lab/Arbora}{\texttt{https://\allowbreak github.com/\allowbreak FutureMLS-Lab/\allowbreak Arbora}}}
\newcommand{\codestatement}{Code is available at \codelink.}
\newcommand{\ackblock}{\section*{Acknowledgments}\aidisclosure}

\title{\texorpdfstring{\includegraphics[height=2.6cm]{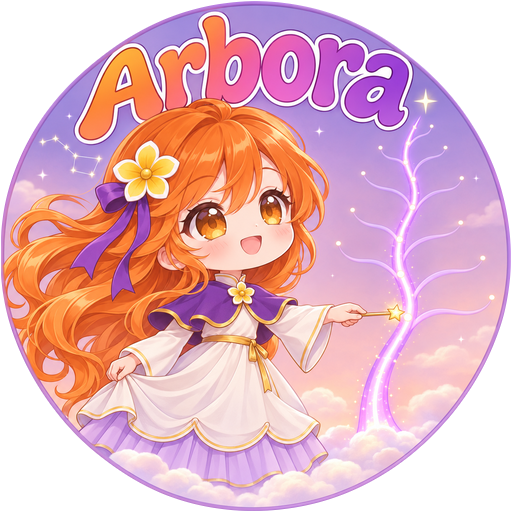}\\[0.5em]}{}\papertitle}

\newcommand{\aff}[1]{\textsuperscript{#1}}
\author{\mdseries
  Yongzhe Wang\aff{1,2,$\dagger$},\enspace
  Fengxiang Bie\aff{1,2,$\dagger$},\enspace
  Yiyang Guo\aff{2,3},\enspace
  Ziyan Chen\aff{1},\enspace
  Cheng Luo, \\
  Zunhai Su\aff{4},\enspace
  Xiaoxia Wu\aff{5},\enspace
  Tianyi Zhang\aff{5},\enspace
  Shuaiwen Leon Song\aff{1},\enspace
  Zhongzhu Zhou\aff{2,$\dagger$} \\[0.7em]
  \normalsize
  \aff{1}University of Sydney \quad
  \aff{2}FutureMLS (Future Machine Learning \& Systems Lab) \\
  \normalsize
  \aff{3}University of California, Santa Cruz \quad
  \aff{4}Tsinghua University \quad
  \aff{5}Together AI
}

\begin{document}
\maketitlewithoverview
\begingroup\renewcommand\thefootnote{}\footnotetext{\raggedright\textsuperscript{$\dagger$}Core contributors. Correspondence to \texttt{zhongzhu.zhou@sydney.edu.au}.}\endgroup
\begin{abstract}
\lineadj{0}{1}%
In speculative decoding, a lightweight drafter proposes several future tokens, and the target model verifies them in one pass.
Block diffusion drafters such as DFlash2 draft a whole block at once, with several candidates per position, but the target verifies only one path through them.
We find that when the target rejects a drafted token, the token it wanted is usually among the candidates but not on this path.
We present \arbora{}, a method that makes better use of these candidates while the trained drafter stays frozen.
Where DFlash2 follows only the best next token, \arbora{}'s path scorer ranks whole paths, and the target verifies a tree of the highest-scoring paths in one pass.
Arbora experts, small networks on the frozen drafter, form a mixture of speculators and learn different preferences where the drafter gives several candidates close scores.
Aggregating their scores re-ranks these candidates, making the correct token more likely to be verified.
Adding an expert also lets \arbora{} learn a new domain online with almost no forgetting.
\arbora{} losslessly raises accept length by \gainALgreedy{} and throughput by \gainTPgreedy{} over DFlash2 on average, across target sizes and architectures, including GLM-5.3-Flash and MiniMax-M3.1-Flash.
\codestatement
\end{abstract}

\section{Introduction}
\label{sec:intro}

\lineadj{1}{1}
Speculative decoding accelerates autoregressive generation in large language models without changing their output distribution \citep{leviathan2023fast,chen2023accelerating}.
The target model generates one token at a time.
A small drafter proposes the next several tokens, and the target model verifies them in one single forward pass.
It verifies them sequentially until one draft token differs from its own distribution.
This takes about as long as generating one token, so several draft tokens can be accepted at the cost of one.
The number of tokens accepted in a pass is the accept length (AL), and a higher AL generally gives a larger speedup.
An autoregressive drafter such as EAGLE-3 \citep{li2025eagle3} runs one forward pass per drafted position, so its cost grows with the draft length.
Block diffusion drafters such as DFlash \citep{chen2026dflash} and DFlash2 \citep{inco2026dflash2} take the target's hidden states as input and draft all positions of a block in a single forward pass, with several candidate tokens at each position.
DFlash outperforms autoregressive drafters such as EAGLE-3 \citep{chen2026dflash}, and DFlash2 improves on it further \citep{inco2026dflash2}.

\lineadj{1}{1}
A block diffusion drafter offers more than a single draft.
It is parallel in two ways: it drafts tokens at all positions of the block at once, and at each position it proposes not one token but several candidates.
For a block of fixed length, the AL then depends on two things: whether the token the target wants is among the candidates at its position and, if it is, whether it is picked for verification.
Not every candidate can be picked, because the target can verify only a limited number of tokens per pass.
DFlash2 follows the best-scoring candidate from one position to the next, so the target verifies a single path through the block.
This raises a natural question:
\emph{At the first token on this path that the target rejects, was the token it wanted missing from the candidates, or among them but not picked for verification?}

\lineadj{1}{1}
We find that it was usually the latter: the drafter had proposed the correct token, but it was not picked for verification.
On Qwen3.8-27B, when DFlash2 stops before the end of the block, in 86\% of cases the correct token is among the candidates (\S\ref{sec:background}).
DFlash2 loses this token because it verifies only one path, and a correct candidate that is not on this path is never verified (Figure~\ref{fig:overview}a).
In the other 14\%, the token was not selected as a candidate at all.
This happens where the drafter is unsure and scores several plausible tokens closely, for example on domains it was not trained on or when the traffic shifts during online training \citep{liu2024osd,qian2026adaflash}.
The candidates therefore need to be chosen better.
Where the drafter is unsure, the token that the target wants also needs a higher rank among them.
We present \arbora{}, a unified method that addresses both problems (Figure~\ref{fig:overview}b).
\arbora{} is lossless, starts from a trained DFlash2 drafter, and keeps the drafter frozen.
Its training follows that of DFlash2 and updates only the path scorer and the Arbora experts.

\lineadj{1}{1}
First, it exploits the dependency between positions with a \textbf{path scorer}.
Like DFlash2's path selector, the path scorer scores each candidate given the token before it, but it ranks whole paths by these scores instead of keeping only the best next token at each position.
The target then verifies a tree of the highest-scoring paths in one forward pass with tree attention \citep{miao2024specinfer}.
Second, it strengthens the parallelism within each position with \textbf{Arbora experts}, small MLPs added to the layers of the frozen drafter.
Together, the Arbora experts on the frozen drafter form a mixture of speculators.
Each Arbora expert gives its own draft of the block, and all of them run in one batched forward pass.
During training, every Arbora expert drafts each block.
We update only the Arbora expert with the most accepted tokens, so the experts learn to prefer different tokens.
Where the drafter is unsure, the Arbora experts can rank the token that the target wants higher among the candidates.

Our contributions are:
\begin{itemize}[leftmargin=1.1em, topsep=3pt, partopsep=0pt, itemsep=2pt, parsep=0pt]
\item \textbf{Observation.} We find that DFlash2 does not make full use of the tokens it drafts: it makes little use of the dependency between positions and has little parallelism within each position (\S\ref{sec:background}).
\item \textbf{Method.} We present \arbora{}, a unified method that addresses both problems.
We train the path scorer directly on the trees it builds, so that it learns to bring the target's own path into the tree.
During training, every Arbora expert drafts each block, and we update only the one with the most tokens accepted by the target.
An added Arbora expert also lets \arbora{} learn a new domain online (\S\ref{sec:arbora}).
\item \textbf{Experiments.} We evaluate \arbora{} on targets of different sizes and architectures (dense, mixture of experts, sparse attention, linear attention, and attention with Gated DeltaNet layers), on public benchmarks spanning code, math, and conversation, under high load, and in online learning (\S\ref{sec:results}).
We integrate \arbora{} into Aurora \citep{wang2026aurora}, so that one framework trains and serves the DFlash2 drafter, the path scorer, and the Arbora experts.
\end{itemize}

On four main targets (Qwen3-8B, Qwen3.8-27B, GLM-5.3-Flash, and MiniMax-M3.1-Flash), \arbora{} losslessly raises the AL and throughput over DFlash2 on public benchmarks from several domains by \gainALgreedy{} and \gainTPgreedy{} on average at temperature $T=0$ and by \gainALsample{} and \gainTPsample{} at $T=1$ (\S\ref{sec:results-main}).
It stays ahead of DFlash2 at high load, gains most on high-entropy tasks, and learns a new domain online with almost no forgetting (\S\ref{sec:results}).

\section{Observation}
\label{sec:background}

We find that DFlash2 makes little use of the dependency between positions and has little parallelism within each position.

\paragraph{Drafting and verification.}
\lineadj{1}{1}%
Each block has $B$ positions ($B=16$ in our setting).
The first position holds the anchor, the last token that the target generated, and the drafter predicts the tokens at the other $B{-}1$ positions.
At each position, its top-$K$ tokens are the candidates \citep{chen2026dflash,inco2026dflash2}.
A path is a sequence of candidates that starts right after the anchor and can end at any position.
DFlash2 picks one candidate at each position, and the target verifies the resulting path in one forward pass \citep{inco2026dflash2}.
The target accepts draft tokens as long as each is the token that it would generate itself, which we call the correct token.
At the first draft token that is not correct, the target generates the correct token itself as the bonus token, which also counts toward the AL.
With tree attention, the target can instead verify several paths in one forward pass \citep{miao2024specinfer,ringel2026ddtree}.
These paths form a tree, and we write $N$ for the number of tokens that the target verifies per step, the anchor and $N{-}1$ draft tokens.
We set $N=16$ in our experiments and vary it in \S\ref{sec:results}.

\paragraph{Where acceptance stops.}
\headgap
\lineadj{1}{0}%
On Qwen3.8-27B, DFlash2 drafts 15 tokens in each step, but the target accepts only 4.03 of them on average\appref{ (Table~\ref{tab:fixed500})}.
When the target rejects a draft token, the correct token is either among the candidates but not on the verified path, or not selected as a candidate.
In trace-driven replay of recorded candidates and target outputs with $N=16$, the first case accounts for 86\% of the rejections and the second for 14\%\appref{ (Appendix~\ref{app:offline})}.
If each step verified the target's own tokens for as long as they are among the candidates, the AL in this replay would be 9.14, against 5.07 for DFlash2.
The candidates thus hold far more accepted tokens than one path can reach.

\paragraph{Little use of the dependency between positions.}
The drafter predicts all positions in one forward pass, so its probabilities at a position do not depend on the candidate chosen before it.
Whether the target accepts a token does depend on the tokens before it.
DFlash2 \citep{inco2026dflash2} therefore adds a path selector that scores every pair of candidates at adjacent positions\appref{ (Eq.~\ref{eq:selector})}.
These scores carry the dependency between positions, but DFlash2 makes little use of them.
It uses them only to pick the best next token at each position, so a correct candidate that is not on its path is never verified.
With the first 4, 8, or 16 candidates, the correct token is among them in 58\%, 75\%, and 86\% of the rejections, but the AL stays at 5.03, 5.07, and 5.07.

\paragraph{Little parallelism within each position.}
\hfill\break%
DFlash2 drafts each block once, so the candidates at each position have only one ranking.
Where the drafter gives several tokens close scores, the correct token can rank too low to be verified, or it is not selected as a candidate at all.
Only a different draft can rank it higher.
Training several complete drafters separately \citep{miao2024specinfer} is a costly way to get different drafts, because each of them adds a full forward pass.
On Qwen3-8B, five such drafters together raise the AL by only 1.7\%, and their additional forward passes slow decoding down more than the higher AL speeds it up\appref{ (Appendix~\ref{app:experts-design})}.

\section{\arbora{}}
\label{sec:arbora}

\begin{figure*}[t]
\centering
\includegraphics[width=\textwidth]{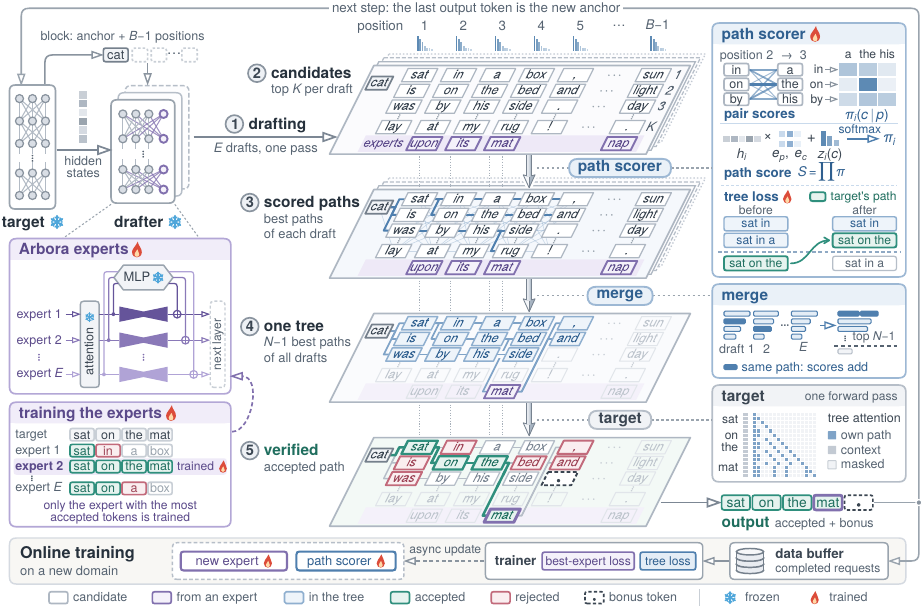}
\caption{Overview of \arbora{}. The center shows one decoding step in five numbered steps, from drafting to the verified output. On the right, the path scorer multiplies pair scores into path scores and is trained with the tree loss, which brings the target's path into the tree. The merge adds the scores of a path that several drafts list, and the target verifies the tree with tree attention. On the left, every draft adds its own Arbora expert to the frozen MLP of each drafter layer, and only the Arbora expert with the most accepted tokens is trained. At the bottom, a new Arbora expert and the path scorer are trained online on completed requests.}
\label{fig:method}
\end{figure*}

Figure~\ref{fig:method} gives an overview of \arbora{}, which starts from a trained DFlash2 drafter and its path selector.
The path scorer reads the hidden states that the drafting pass already computes, so neither the path scorer nor the Arbora experts add a forward pass of the drafter.

\ifpreprintlayout\begin{algorithm}[t]
\small
\caption{Selecting the tree to verify.}
\label{alg:tree}
\begin{algorithmic}[1]
\Require anchor $x_0$; candidates $C_1,\dots,C_{B-1}$; pair scores $\pi_i(c \mid p)$; $N$ tokens verified per step
\Ensure the $N{-}1$ paths of the tree
\State $\mathcal{P}_0 \gets \{x_0\}$, \ $S(x_0) \gets 1$ \Comment{the empty path}
\For{$i = 1,\dots,B{-}1$}
  \For{each $u \in \mathcal{P}_{i-1}$ ending in $p$, and each $c \in C_i$}
    \State $S(uc) \gets S(u)\cdot \pi_i(c \mid p)$ \Comment{extend $u$ by $c$}
  \EndFor
  \State $\mathcal{P}_i \gets$ the $N{-}1$ paths $uc$ with the highest $S$
\EndFor
\State \Return the $N{-}1$ paths with the highest $S$ in $\bigcup_i \mathcal{P}_i$
\end{algorithmic}
\end{algorithm}
\fi

\subsection{Scoring paths and selecting the tree}
\label{sec:paths}

Like DFlash2's path selector, the path scorer scores every pair of candidates at adjacent positions (Figure~\ref{fig:method}, path scorer).
For a candidate $p$ at position $i{-}1$ and a candidate $c$ at position $i$, the score is the drafter's logit of $c$ plus a term computed from the hidden state at position $i$ and the embeddings of $p$ and $c$\appref{ (Eq.~\ref{eq:selector} in Appendix~\ref{app:formulas})}.
The scores of all $K \times K$ pairs of a position are computed at once.
DFlash2 uses them only to pick the best next candidate.
The path scorer normalizes them over the $K$ candidates of position $i$, so the pair score $\pi_i(c \mid p)$ is the probability of $c$ at position $i$ given $p$ at position $i{-}1$\appref{ (Eq.~\ref{eq:node})}.
For $i=1$, $p$ is the anchor.
The score of a path $y_{1:k}$ is the product of the pair scores along it,
\begin{equation}
S(y_{1:k}) = \prod_{i=1}^{k} \pi_i(y_i \mid y_{i-1}),
\label{eq:pathscore}
\end{equation}
where $y_0$ is the anchor.
It estimates the probability that the target accepts every token on the path.
When the target verifies $N$ tokens per step, the $N{-}1$ paths with the highest such probability give the highest expected AL\appref{ (Eq.~\ref{eq:yield})} \citep{li2024eagle2,wang2025opttree}.
In Figure~\ref{fig:overview}b, ``the'' scores .78 after ``on'', so ``sat on the'' scores .26 and enters the five best paths.

\paragraph{Selecting the tree.}
\label{sec:verify}
Algorithm~\ref{alg:tree} selects the $N{-}1$ paths with the highest scores for verification.
Every pair score is a probability and is at most 1, so a path never scores higher than any of its prefixes (Eq.~\ref{eq:pathscore}).
When a path is among the $N{-}1$ highest-scoring paths, its prefixes are therefore among them too, and these paths form a tree.
Where the drafter is confident and one candidate takes most of the probability, the tree grows deep.
Where several candidates have similar probabilities, it branches.
Algorithm~\ref{alg:tree} runs on the GPU with a fixed number of batched sorts.
At each position, it extends every kept path by every candidate and keeps the $N{-}1$ extensions with the highest scores.
It then takes the $N{-}1$ highest-scoring paths among all that it kept as the tree.
Keeping $N{-}1$ paths at each position is enough, because the whole tree has only $N{-}1$ paths.

\subsection{Training the path scorer on its own trees}
\label{sec:treeloss}

DFlash2 trains its path selector with a cross-entropy at each position.
Given the target's previous token, this loss raises the score of the target's token among the candidates of that position.
It compares only the candidates of one position, so it does not teach the path selector which paths should enter a tree.
In \arbora{}, a path enters the tree only if its score is among the $N{-}1$ highest, so it competes with all other paths, including paths with a different prefix or a different length.
\arbora{} therefore trains the path scorer on the trees it builds (Figure~\ref{fig:method}, tree loss).
\ifpreprintlayout\else\begin{algorithm}[t]
\small
\caption{Selecting the tree to verify.}
\label{alg:tree}
\begin{algorithmic}[1]
\Require anchor $x_0$; candidates $C_1,\dots,C_{B-1}$; pair scores $\pi_i(c \mid p)$; $N$ tokens verified per step
\Ensure the $N{-}1$ paths of the tree
\State $\mathcal{P}_0 \gets \{x_0\}$, \ $S(x_0) \gets 1$ \Comment{the empty path}
\For{$i = 1,\dots,B{-}1$}
  \For{each $u \in \mathcal{P}_{i-1}$ ending in $p$, and each $c \in C_i$}
    \State $S(uc) \gets S(u)\cdot \pi_i(c \mid p)$ \Comment{extend $u$ by $c$}
  \EndFor
  \State $\mathcal{P}_i \gets$ the $N{-}1$ paths $uc$ with the highest $S$
\EndFor
\State \Return the $N{-}1$ paths with the highest $S$ in $\bigcup_i \mathcal{P}_i$
\end{algorithmic}
\end{algorithm}
\fi
In training, the tokens that the target generated after the anchor are known, and we call this sequence the target's path.
For each block, we build the tree in the same way as at inference and train the path scorer so that the target's path is in it.
Let $x_{1:m}$ be the longest prefix of the target's path whose tokens are all candidates.
The $k{-}1$ shorter prefixes of a prefix $x_{1:k}$ hold $k{-}1$ of the $N{-}1$ places in the tree.
The prefix $x_{1:k}$ is therefore in the tree when its score exceeds the threshold $\tau_k$, the $(N{-}k)$-th highest score among the paths that leave the target's path at position $k$ or earlier\appref{ (Appendix~\ref{app:treeloss})}.
The tree loss trains every prefix of length at most $N{-}1$ to exceed its threshold.
With $\Delta_k = \log S(x_{1:k}) - \log \tau_k$, it is
\begin{equation}
L_{\mathrm{tree}} = -\sum_{k=1}^{\min(m,\,N-1)} \log \sigma(\beta\,\Delta_k),
\label{eq:treeloss}
\end{equation}
where $\sigma$ is the logistic function and $\beta$ is a scaling constant.
The loss is almost zero for a prefix well inside the tree.
For a prefix that is outside the tree or close to its threshold, the loss raises the prefix's score and lowers the score of the path at the threshold.
Choosing the paths is not differentiable, so training treats the chosen paths as fixed.
Their scores are products of pair scores, and the gradient reaches the pair scores through these products.
The path scorer starts from the weights of DFlash2's path selector, and we train it with $L_{\mathrm{tree}} + \lambda L_{\mathrm{CE}}$, where $L_{\mathrm{CE}}$ is DFlash2's cross-entropy\appref{ (Algorithm~\ref{alg:treeloss} and the values of $\beta$ and $\lambda$ in Appendix~\ref{app:treeloss})}.

\subsection{Arbora experts}
\label{sec:experts}

Arbora experts strengthen the parallelism within each position.
We call the frozen drafter the shared drafter, because every draft uses its weights.
The Arbora experts on the shared drafter form the mixture of speculators.

\paragraph{Architecture.}
In every drafter layer $\ell$, $E$ Arbora experts $f^{(1)}_\ell,\dots,f^{(E)}_\ell$ are added beside the frozen MLP of the shared drafter (Figure~\ref{fig:method}, Arbora experts).
Each is an MLP whose hidden dimension is 1/8 of that of the frozen MLP.
Draft $e$ uses Arbora expert $e$ in every layer and at every position, and no router chooses among the Arbora experts.
The Arbora expert's output is added to that of the frozen MLP,
\begin{equation}
\mathrm{MLP}^{(e)}_\ell(u) = \mathrm{MLP}_\ell(u) + f^{(e)}_\ell(u),
\label{eq:expert}
\end{equation}
where $u$ is the MLP input in draft $e$.
The $E$ drafts share all other weights of the shared drafter and run in one batched forward pass.
Eight Arbora experts add 0.26\,ms to an 8.62\,ms step on Qwen3-8B, less than the 0.95\,ms that the attention and MLP of one more complete drafter take\appref{ (Appendix~\ref{app:latency})}.
In serving, the number of Arbora experts can be chosen according to the load.

\begin{table*}[t]
\centering
\scriptsize
\setlength{\tabcolsep}{0pt}
\definecolor{tonecell0}{HTML}{EEF2F7}\definecolor{toneband0}{HTML}{D7E2F0}\definecolor{tonecell1}{HTML}{F1EEF8}\definecolor{toneband1}{HTML}{DCD4EE}
\setlength{\aboverulesep}{0pt}\setlength{\belowrulesep}{0pt}
\setbox\strutbox\hbox{\vrule height6.35pt depth1.65pt width0pt}
\begin{tabular}{@{}wl{39.0pt}@{\hspace{5.0pt}}wr{13.5pt}@{}wr{27.5pt}@{\hspace{11.25pt}}>{\columncolor{white}[3.25pt][0pt]}wr{13.5pt}@{}>{\columncolor{white}[0pt][4.5pt]}wr{27.5pt}@{\hspace{11.248571428571429pt}}wr{13.5pt}@{}wr{27.5pt}@{\hspace{11.25pt}}>{\columncolor{white}[3.25pt][0pt]}wr{13.5pt}@{}>{\columncolor{white}[0pt][4.5pt]}wr{27.5pt}@{\hspace{11.248571428571429pt}}wr{13.5pt}@{}wr{27.5pt}@{\hspace{11.25pt}}>{\columncolor{white}[3.25pt][0pt]}wr{13.5pt}@{}>{\columncolor{white}[0pt][4.5pt]}wr{27.5pt}@{\hspace{11.248571428571429pt}}wr{13.5pt}@{}wr{27.5pt}@{\hspace{11.25pt}}>{\columncolor{white}[3.25pt][0pt]}wr{13.5pt}@{}>{\columncolor{white}[0pt][4.5pt]}wr{27.5pt}@{\hspace{4.5pt}}}
 & \multicolumn{4}{c@{\hspace{11.248571428571429pt}}}{Qwen3-8B} & \multicolumn{4}{c@{\hspace{11.248571428571429pt}}}{Qwen3.8-27B} & \multicolumn{4}{c@{\hspace{11.248571428571429pt}}}{GLM-5.3-Flash} & \multicolumn{4}{c@{\hspace{4.5pt}}}{MiniMax-M3.1-Flash} \\
\cmidrule(l{1.25pt}r{6.75pt}){2-5}\cmidrule(l{1.25pt}r{6.75pt}){6-9}\cmidrule(l{1.25pt}r{6.75pt}){10-13}\cmidrule(l{1.25pt}r{0.00pt}){14-17}
\rule{0pt}{8.2pt} & \multicolumn{2}{c@{\hspace{11.25pt}}}{DFlash2} & \multicolumn{2}{c@{\hspace{11.248571428571429pt}}}{\arbora{}} & \multicolumn{2}{c@{\hspace{11.25pt}}}{DFlash2} & \multicolumn{2}{c@{\hspace{11.248571428571429pt}}}{\arbora{}} & \multicolumn{2}{c@{\hspace{11.25pt}}}{DFlash2} & \multicolumn{2}{c@{\hspace{11.248571428571429pt}}}{\arbora{}} & \multicolumn{2}{c@{\hspace{11.25pt}}}{DFlash2} & \multicolumn{2}{c@{\hspace{4.5pt}}}{\arbora{}} \\
\cmidrule(l{1pt}r{11.25pt}){2-3}\cmidrule(l{1pt}r{11.25pt}){4-5}\cmidrule(l{1pt}r{11.25pt}){6-7}\cmidrule(l{1pt}r{11.25pt}){8-9}\cmidrule(l{1pt}r{11.25pt}){10-11}\cmidrule(l{1pt}r{11.25pt}){12-13}\cmidrule(l{1pt}r{11.25pt}){14-15}\cmidrule(l{1pt}r{4.50pt}){16-17}
\rule{0pt}{8.2pt}Benchmark & AL & tok/s & AL & tok/s & AL & tok/s & AL & tok/s & AL & tok/s & AL & tok/s & AL & tok/s & AL & tok/s \\[1pt]
\midrule
\multicolumn{17}{@{}l}{\rule{0pt}{8.2pt}\textit{$T=0$}} \\
HumanEval & 5.48 & 629 & \cellcolor{tonecell0}\textbf{6.66} & \cellcolor{tonecell0}\textbf{731} & 5.07 & 300 & \cellcolor{tonecell0}\textbf{6.42} & \cellcolor{tonecell0}\textbf{371} & 6.55 & 563 & \cellcolor{tonecell0}\textbf{7.93} & \cellcolor{tonecell0}\textbf{676} & 5.47 & 532 & \cellcolor{tonecell0}\textbf{7.05} & \cellcolor{tonecell0}\textbf{658} \\
MBPP & 5.49 & 630 & \cellcolor{tonecell0}\textbf{6.64} & \cellcolor{tonecell0}\textbf{730} & 5.66 & 331 & \cellcolor{tonecell0}\textbf{7.08} & \cellcolor{tonecell0}\textbf{406} & 7.05 & 606 & \cellcolor{tonecell0}\textbf{8.46} & \cellcolor{tonecell0}\textbf{719} & 5.39 & 524 & \cellcolor{tonecell0}\textbf{6.83} & \cellcolor{tonecell0}\textbf{640} \\
GSM8K & 6.72 & 767 & \cellcolor{tonecell0}\textbf{7.81} & \cellcolor{tonecell0}\textbf{854} & 6.63 & 375 & \cellcolor{tonecell0}\textbf{7.98} & \cellcolor{tonecell0}\textbf{441} & 6.80 & 572 & \cellcolor{tonecell0}\textbf{8.43} & \cellcolor{tonecell0}\textbf{693} & 5.37 & 504 & \cellcolor{tonecell0}\textbf{6.95} & \cellcolor{tonecell0}\textbf{613} \\
MATH-500 & 6.35 & 728 & \cellcolor{tonecell0}\textbf{7.51} & \cellcolor{tonecell0}\textbf{823} & 6.05 & 351 & \cellcolor{tonecell0}\textbf{7.34} & \cellcolor{tonecell0}\textbf{417} & 7.19 & 615 & \cellcolor{tonecell0}\textbf{8.70} & \cellcolor{tonecell0}\textbf{732} & 5.82 & 571 & \cellcolor{tonecell0}\textbf{7.37} & \cellcolor{tonecell0}\textbf{685} \\
MT-Bench 1 & 4.19 & 486 & \cellcolor{tonecell0}\textbf{5.34} & \cellcolor{tonecell0}\textbf{591} & 4.04 & 239 & \cellcolor{tonecell0}\textbf{5.29} & \cellcolor{tonecell0}\textbf{306} & 3.77 & 343 & \cellcolor{tonecell0}\textbf{4.83} & \cellcolor{tonecell0}\textbf{438} & 3.14 & 326 & \cellcolor{tonecell0}\textbf{4.16} & \cellcolor{tonecell0}\textbf{413} \\
MT-Bench 2 & 4.09 & 469 & \cellcolor{tonecell0}\textbf{5.35} & \cellcolor{tonecell0}\textbf{582} & 3.72 & 220 & \cellcolor{tonecell0}\textbf{4.94} & \cellcolor{tonecell0}\textbf{286} & 3.71 & 331 & \cellcolor{tonecell0}\textbf{4.82} & \cellcolor{tonecell0}\textbf{427} & 3.27 & 330 & \cellcolor{tonecell0}\textbf{4.30} & \cellcolor{tonecell0}\textbf{416} \\
\noalign{\vskip\dimexpr 1.6pt+\cmidrulewidth\relax}
\noalign{\vbox to 0pt{\hbox{\textcolor{toneband0}{\rule[-10.85pt]{455.24pt}{10.85pt}}}\vss}}
\rule{0pt}{8.2pt}\textbf{Mean gain} &  &  & \cellcolor{toneband0} & \cellcolor{toneband0}\makebox[0pt][r]{\makebox[48.75pt][r]{\textbf{+22.5\%}\hspace{3.5pt}\textbf{+17.0\%}}\kern-3.0pt} &  &  & \cellcolor{toneband0} & \cellcolor{toneband0}\makebox[0pt][r]{\makebox[48.75pt][r]{\textbf{+26.2\%}\hspace{3.5pt}\textbf{+23.5\%}}\kern-3.0pt} &  &  & \cellcolor{toneband0} & \cellcolor{toneband0}\makebox[0pt][r]{\makebox[48.75pt][r]{\textbf{+24.0\%}\hspace{3.5pt}\textbf{+22.6\%}}\kern-3.0pt} &  &  & \cellcolor{toneband0} & \cellcolor{toneband0}\makebox[0pt][r]{\makebox[48.75pt][r]{\textbf{+29.3\%}\hspace{3.5pt}\textbf{+23.3\%}}\kern-3.0pt} \\[1pt]
\midrule
\multicolumn{17}{@{}l}{\rule{0pt}{8.2pt}\textit{$T=1$}} \\
HumanEval & 4.93 & 563 & \cellcolor{tonecell1}\textbf{6.16} & \cellcolor{tonecell1}\textbf{672} & 3.56 & 214 & \cellcolor{tonecell1}\textbf{4.66} & \cellcolor{tonecell1}\textbf{273} & 5.78 & 500 & \cellcolor{tonecell1}\textbf{7.12} & \cellcolor{tonecell1}\textbf{607} & 4.36 & 428 & \cellcolor{tonecell1}\textbf{5.75} & \cellcolor{tonecell1}\textbf{545} \\
MBPP & 4.78 & 547 & \cellcolor{tonecell1}\textbf{5.99} & \cellcolor{tonecell1}\textbf{654} & 3.82 & 226 & \cellcolor{tonecell1}\textbf{4.90} & \cellcolor{tonecell1}\textbf{284} & 6.03 & 521 & \cellcolor{tonecell1}\textbf{7.39} & \cellcolor{tonecell1}\textbf{632} & 4.45 & 434 & \cellcolor{tonecell1}\textbf{5.77} & \cellcolor{tonecell1}\textbf{542} \\
GSM8K & 5.63 & 641 & \cellcolor{tonecell1}\textbf{6.81} & \cellcolor{tonecell1}\textbf{741} & 4.93 & 283 & \cellcolor{tonecell1}\textbf{6.06} & \cellcolor{tonecell1}\textbf{340} & 6.18 & 522 & \cellcolor{tonecell1}\textbf{7.72} & \cellcolor{tonecell1}\textbf{637} & 4.52 & 429 & \cellcolor{tonecell1}\textbf{5.88} & \cellcolor{tonecell1}\textbf{527} \\
MATH-500 & 5.54 & 630 & \cellcolor{tonecell1}\textbf{6.73} & \cellcolor{tonecell1}\textbf{734} & 4.33 & 253 & \cellcolor{tonecell1}\textbf{5.41} & \cellcolor{tonecell1}\textbf{310} & 6.24 & 535 & \cellcolor{tonecell1}\textbf{7.63} & \cellcolor{tonecell1}\textbf{647} & 4.91 & 484 & \cellcolor{tonecell1}\textbf{6.32} & \cellcolor{tonecell1}\textbf{589} \\
MT-Bench 1 & 3.57 & 413 & \cellcolor{tonecell1}\textbf{4.70} & \cellcolor{tonecell1}\textbf{524} & 2.97 & 175 & \cellcolor{tonecell1}\textbf{3.82} & \cellcolor{tonecell1}\textbf{223} & 3.32 & 301 & \cellcolor{tonecell1}\textbf{4.34} & \cellcolor{tonecell1}\textbf{393} & 2.70 & 281 & \cellcolor{tonecell1}\textbf{3.61} & \cellcolor{tonecell1}\textbf{357} \\
MT-Bench 2 & 3.48 & 398 & \cellcolor{tonecell1}\textbf{4.58} & \cellcolor{tonecell1}\textbf{501} & 2.84 & 170 & \cellcolor{tonecell1}\textbf{3.71} & \cellcolor{tonecell1}\textbf{216} & 3.25 & 289 & \cellcolor{tonecell1}\textbf{4.24} & \cellcolor{tonecell1}\textbf{382} & 2.76 & 278 & \cellcolor{tonecell1}\textbf{3.63} & \cellcolor{tonecell1}\textbf{352} \\
\noalign{\vskip\dimexpr 1.6pt+\cmidrulewidth\relax}
\noalign{\vbox to 0pt{\hbox{\textcolor{toneband1}{\rule[-10.85pt]{455.24pt}{10.85pt}}}\vss}}
\rule{0pt}{8.2pt}\textbf{Mean gain} &  &  & \cellcolor{toneband1} & \cellcolor{toneband1}\makebox[0pt][r]{\makebox[48.75pt][r]{\textbf{+26.1\%}\hspace{3.5pt}\textbf{+20.7\%}}\kern-3.0pt} &  &  & \cellcolor{toneband1} & \cellcolor{toneband1}\makebox[0pt][r]{\makebox[48.75pt][r]{\textbf{+27.8\%}\hspace{3.5pt}\textbf{+25.2\%}}\kern-3.0pt} &  &  & \cellcolor{toneband1} & \cellcolor{toneband1}\makebox[0pt][r]{\makebox[48.75pt][r]{\textbf{+25.7\%}\hspace{3.5pt}\textbf{+24.7\%}}\kern-3.0pt} &  &  & \cellcolor{toneband1} & \cellcolor{toneband1}\makebox[0pt][r]{\makebox[48.75pt][r]{\textbf{+30.9\%}\hspace{3.5pt}\textbf{+25.1\%}}\kern-3.0pt} \\[1pt]
\midrule
\end{tabular}
\caption{AL and throughput (tok/s) of DFlash2 and \arbora{} on public benchmarks at $T=0$ and $T=1$. MT-Bench~1 and~2 are the first and second turn. Both methods use the same drafter, which \arbora{} keeps frozen. Bold marks the higher of the two methods. Mean gain is \arbora{} over DFlash2, averaged over the six rows.}
\label{tab:main}
\end{table*}

\paragraph{Merging the drafts.}
Each draft has its own hidden states and its own candidates.
The path scorer computes the pair scores of a draft from that draft's hidden states, and Algorithm~\ref{alg:tree} selects the draft's $N{-}1$ highest-scoring paths (Figure~\ref{fig:method}, steps~3 and~4).
When several drafts select the same path, the merge keeps it once and adds its scores.
The $N{-}1$ merged paths with the highest scores form the tree that the target verifies\appref{ (Appendix~\ref{app:merge})}.

\paragraph{Training.}
\headgap
Training follows multiple choice learning \citep{guzmanrivera2012multiple,lee2016stochastic}, which updates only the best predictor for each example.
Every Arbora expert drafts each training block without gradients, and DFlash2's path selector turns each draft into one path.
A token on this path counts as accepted if it and every token before it equal the target's output.
Only the Arbora expert with the most accepted tokens is trained on the block, with a cross-entropy on the target's tokens that weights earlier positions more (Figure~\ref{fig:method}, training the experts\appref{; Eq.~\ref{eq:mcl} in Appendix~\ref{app:formulas}}).
As a result, the Arbora experts differ from the shared drafter mostly where it is unsure.
We measure how unsure the shared drafter is at a position by the ratio $p_2/p_1$ of the probabilities of its second and first tokens.
On Qwen3-8B, when the two are close ($p_2/p_1 \ge 0.75$), the top token of an Arbora expert differs from that of the shared drafter at 60\% of positions.
When the shared drafter is confident ($p_2/p_1 < 0.01$), it differs at only 3\%\appref{ (Appendix~\ref{app:training})}.

\paragraph{New domains.}
For a new domain, one Arbora expert is added and trained online, while the shared drafter and the existing Arbora experts stay frozen (Figure~\ref{fig:method}, bottom; \S\ref{sec:online}).
The server keeps completed requests in a data buffer.
A trainer trains the new Arbora expert on them with the loss above, the best-expert loss in Figure~\ref{fig:method}, and can also train the path scorer with the tree loss.
The server loads the updated weights asynchronously while it keeps serving.

\section{Experiments}
\label{sec:results}

\subsection{Setup}
\label{sec:setup}

\paragraph{Targets.}
We evaluate four main targets: Qwen3-8B \citep{yang2025qwen3} (dense), Qwen3.8-27B (attention with Gated DeltaNet layers), GLM-5.3-Flash (mixture of experts with linear and sparse attention layers), and MiniMax-M3.1-Flash (mixture of experts with sparse attention)\appref{; the other targets we tested are in Appendix~\ref{app:more-targets}}.

\paragraph{Methods.}
We compare \arbora{} with DFlash2 on the same drafter, with $K=16$ candidates per position (32 on MiniMax-M3.1-Flash, the setting of its released drafter) and $N=16$ tokens verified per step.
\arbora{} runs the path scorer with $E=8$ Arbora experts\appref{ (Appendix~\ref{app:training})}.

\begin{figure}[t]
\centering
\includegraphics[width=\columnwidth]{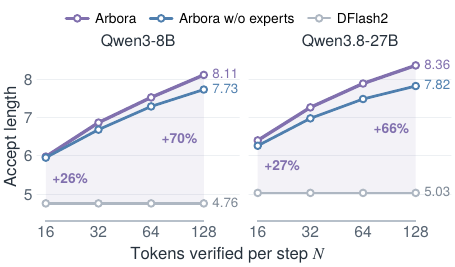}
\caption{AL against the number $N$ of tokens verified per step, on fixed-500 at $T=0$.}
\label{fig:nscan}
\vspace{-6pt}%
\end{figure}

\paragraph{Evaluation.}
The main table uses five public benchmarks: HumanEval \citep{chen2021evaluating} and MBPP \citep{austin2021program} for code, GSM8K \citep{cobbe2021training} and MATH-500 \citep{hendrycks2021measuring,lightman2024lets} for math, and MT-Bench \citep{zheng2023judging} for conversation.
On these benchmarks, we decode at temperature $T=0$ and $T=1$, one request at a time, with up to 512 generated tokens and \textbf{thinking mode enabled}.
The ablations, the scans over the number of verified tokens, and the tests at high load use fixed-500, a set of prompts from five domains (code, math, factual questions, creative writing, and chat) that is held out from all training data.

\paragraph{Framework.}
We integrate DFlash2 and \arbora{} into Aurora \citep{wang2026aurora}, which trains them and serves them on SGLang \citep{zheng2024sglang}\appref{ (parameter settings in Appendix~\ref{app:rounds})}.
With it we train the whole Qwen3-8B DFlash2 drafter.
On every target, \arbora{} keeps the DFlash2 drafter frozen and trains only the path scorer and the Arbora experts.
Qwen3.8-27B, GLM-5.3-Flash, and MiniMax-M3.1-Flash build on released DFlash2 drafters\appref{ (Appendix~\ref{app:training})}.

\subsection{Main results}
\label{sec:results-main}

\arbora{} has a higher AL and throughput than DFlash2 on every benchmark, on every main target, and at both $T=0$ and $T=1$ (Table~\ref{tab:main}).
Averaged over the four targets and the six rows, it raises the AL and throughput by \gainALgreedy{} and \gainTPgreedy{} at $T=0$ and by \gainALsample{} and \gainTPsample{} at $T=1$.
On fixed-500, whose prompts follow the distribution of the training data, the gains are 28.3\% and 22.5\% at $T=0$ and 28.9\% and 23.5\% at $T=1$\appref{ (Appendix~\ref{app:fixed500})}.

\arbora{} also has a higher AL than other tree methods on the same drafter.
At $T=0$ on fixed-500, it exceeds TAPS \citep{wang2026taps} by 3.6--8.2\%, GRAFT \citep{ye2026graft} by 5.1--14.9\%, and DDTree \citep{ringel2026ddtree} by 12.6--16.0\% on the four main targets\appref{ (Appendix~\ref{app:tree-methods})}.

The AL gain is largest on high-entropy tasks such as conversation and creative writing, where the target's most probable token has a low probability\appref{ (Appendix~\ref{app:public})}.

The gain also grows with the number $N$ of tokens verified per step (Figure~\ref{fig:nscan}).
DFlash2 verifies a single path of at most 15 draft tokens, so its AL stays the same at larger $N$.
From $N=16$ to $N=128$, \arbora{}'s AL gain over DFlash2 rises from 26\% to 70\% on Qwen3-8B and from 27\% to 66\% on Qwen3.8-27B\appref{ (Appendix~\ref{app:budget})}.

\subsection{Ablation}
\label{sec:results-analysis}

We also run the path scorer and the Arbora experts each on their own\appref{ (Appendix~\ref{app:ablation})}.
On fixed-500 at $T=0$, each raises the AL over DFlash2 by more than 22\% on average, and together they raise it by 28.3\%.
The two parts also complement each other: the gain of the Arbora experts grows with $N$ (Figure~\ref{fig:nscan}), and adding an Arbora expert lets \arbora{} learn a new domain online (\S\ref{sec:online}).

\paragraph{Path scorer.}
Both the tree and the tree loss contribute to the path scorer's gain.
With the weights of DFlash2's path selector, its tree already raises the throughput over DFlash2 by 18.7\% on average, and training with the tree loss raises this to 21.6\%\appref{ (Table~\ref{tab:selector})}.

\paragraph{Arbora experts.}
The gain of the Arbora experts does not come from added parameters.
On Qwen3-8B, eight Arbora experts raise the trace-driven AL by 10.2\% over the shared drafter alone, while one MLP with the same number of parameters leaves it unchanged\appref{ (Appendix~\ref{app:experts-design})}.
Training a random Arbora expert on each block instead of the best one also gains nothing.
We use eight Arbora experts of 1/8 width, the setting that worked best in our tests, and other targets or hardware may favor a different number or width.
The gain comes from the candidates that the Arbora experts bring into the tree.
With $N=16$ tokens verified per step, their tree holds 17\% more distinct candidates than the tree of the shared drafter, and with $N=64$ it holds 33\% more.
At $N=16$, most of the gain comes from tokens that the shared drafter ranks in its top 8, and at $N=64$ a larger share comes from tokens below them\appref{ (Table~\ref{tab:experts-topk})}.
The AL gain of the Arbora experts therefore grows with $N$.
On Qwen3.8-27B, they add 2.2\% at $N=16$, 5.4\% at $N=64$, and 6.9\% at $N=128$\appref{ (Table~\ref{tab:nscan})}.

\begin{figure}[t]
\centering
\includegraphics[width=\columnwidth]{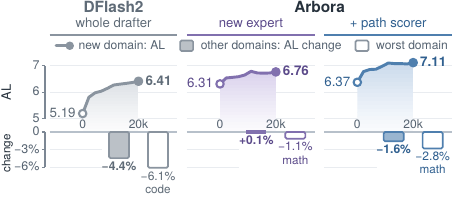}
\caption{Learning a new domain online on Qwen3-8B over a stream of 20{,}000 requests. Above the axis: AL on the stream of the new domain, per 2{,}000 requests. Below the axis: the change of AL on the five other domains after the stream. The filled bar is the five domains together, and the open bar is the domain that loses most. ``+ path scorer'' trains the new Arbora expert and the path scorer.}
\label{fig:online-main}
\end{figure}

\subsection{High load}
\label{sec:results-concurrency}

At high load the GPU is compute-bound, so the number of Arbora experts can be reduced according to the hardware and the system load.
\arbora{} stays ahead of DFlash2 at every measured load.
With 4 to 32 requests, it is 13.9--15.2\% faster on Qwen3-8B, and with 4 to 16 requests, it is 13.6--16.6\% faster on Qwen3.8-27B\appref{ (Appendix~\ref{app:concurrency})}.

\subsection{Learning a new domain online}
\label{sec:online}

\arbora{} learns a new domain with the online loop of \S\ref{sec:experts}, which follows Aurora \citep{wang2026aurora}.
It adds one Arbora expert and trains either this Arbora expert alone or it together with the path scorer.
DFlash2 is trained online on the same stream from the same drafter and updates its whole drafter at its best learning rate.
We serve 20{,}000 software-engineering requests from SWE-smith \citep{yang2025swesmith} on Qwen3-8B.
After the stream, we test on 500 held-out SWE-bench Verified turns \citep{jimenez2024swebench,chowdhury2024verified} for the new domain and on fixed-500, whose five domains are not trained online.

Both \arbora{} settings have a higher AL than DFlash2 throughout the stream and forget less (Figure~\ref{fig:online-main}).
Training only the new Arbora expert leaves the other domains almost unchanged.
Training the path scorer as well raises the AL on the new domain further, at a small loss on the others.
DFlash2, which updates its whole drafter, loses the most on the other domains.
On the held-out SWE-bench Verified turns, the AL after the stream is 6.54 and 6.84 for the two \arbora{} settings, against 5.99 for DFlash2\appref{ (Appendix~\ref{app:online})}.

\section{Related Work}
\label{sec:related}

\paragraph{Drafters.}
DFlash \citep{chen2026dflash} drafts a block in one forward pass of a small block diffusion model.
Domino \citep{huang2026domino}, DSpark \citep{cheng2026dspark}, XPress \citep{wang2026xpress}, and DSpine \citep{zhang2026dspine} make each position's prediction depend on earlier tokens, and each of them has to train its drafter.
DFlash2 \citep{inco2026dflash2} and LiLiCorr \citep{rusanovsky2026lilicorr} score pairs of adjacent candidates.
All of them verify one path.
\arbora{} keeps the drafter frozen.
It ranks whole paths through the drafter's candidates and verifies a tree of the best paths.
This costs less and gains more.
In our measurements on Qwen3-8B with released weights, DSpine and XPress reach an AL of 3.16 and 3.27, against 2.74 for DFlash2, while \arbora{}'s path scorer reaches 3.44 on the released DFlash2 drafter without any training\appref{ (Appendix~\ref{app:released})}.

\paragraph{Trees.}
EAGLE-2 \citep{li2024eagle2} builds a dynamic draft tree from the paths that the target is most likely to accept, but its autoregressive drafter has to run once for every tree layer.
\arbora{} selects its tree by the same criterion (\S\ref{sec:paths}) from a single drafting pass.
DDTree \citep{ringel2026ddtree} builds a best-first tree from the candidates of a block diffusion drafter, and JetSpec \citep{hu2026jetspec} builds one from the candidates of a parallel draft head.
DDTree scores a path by the probability of each token at its own position, without the token before it.
GRAFT \citep{ye2026graft} and TAPS \citep{wang2026taps} score a token given the token before it, with pair scorers trained on recorded traces.
\arbora{} instead trains its path scorer on the trees it builds, so that the target's path is in the tree.

\paragraph{Several drafts.}
\lineadj{1}{0}%
SpecInfer \citep{miao2024specinfer} merges the token trees of several boost-tuned drafters into one.
In our test, five separately trained drafters merged in this way raise the AL by only 1.7\% (\S\ref{sec:background}).
D$^2$SD \citep{zhang2026d2sd} trains a second complete drafter that redrafts where rejection is most likely, which adds a drafter pass to every step.
\arbora{} instead adds Arbora experts to one frozen drafter, and all drafts run in one batched forward pass.
Unlike sparse expert layers \citep{shazeer2017outrageously,fedus2022switch}, the Arbora experts have no router.
We train them with multiple choice learning \citep{guzmanrivera2012multiple,lee2016stochastic}, so that they prefer different tokens.

\paragraph{Online learning and serving.}
\looseacl
Online speculative decoding \citep{liu2024osd}, Aurora \citep{wang2026aurora}, and AdaFlash \citep{qian2026adaflash} train the drafter on live traffic.
Test-Time Speculation \citep{kumar2026testtime} adapts the drafter within one answer, and Osprey \citep{bie2026osprey} pretrains it on broad data before deployment.
\arbora{} instead keeps the drafter frozen and trains one added Arbora expert online for a new domain.
OSCAR \citep{zhou2026oscar} reduces the cost of a target pass with a 2-bit KV cache and a fused attention kernel.
In an early version, we used its fused attention to share the attention of five separately trained drafters and trained only their MLPs separately, which raised the AL only slightly.

\section{Conclusion}
\label{sec:conclusion}

\lineadj{1}{1}%
We find that DFlash2 makes little use of the dependency between positions and has little parallelism within each position.
When it stops before the end of the block, the correct token is usually among the candidates but not on the one path that it verifies.
\arbora{} is a unified method that addresses both problems while the drafter stays frozen.
Its path scorer ranks whole paths and is trained on the trees it builds, and its Arbora experts form a mixture of speculators that gives several drafts of the block in one batched forward pass.
On public benchmarks, \arbora{} raises the AL and throughput over DFlash2 by \gainALgreedy{} and \gainTPgreedy{} on average at $T=0$ and by \gainALsample{} and \gainTPsample{} at $T=1$.
The gain is largest on high-entropy tasks, and it grows with the number of tokens verified per step.
\arbora{} stays ahead of DFlash2 under load.
It also learns a new domain online by adding one Arbora expert, with almost no forgetting of other domains.

\section*{Limitations}

Our serving implementation optimizes only the parts that \arbora{} adds, so further engineering can raise its throughput.
We measure throughput on H100 GPUs.
With more verified tokens or more Arbora experts, a step becomes more compute-bound, so on faster hardware the same AL gain could give a larger throughput gain.
Under load, we report throughput at fixed numbers of concurrent requests.
Latency under real request arrivals also depends on the serving engine and its scheduler.

\ackblock

\bibliography{custom}

\appendix
\setcounter{topnumber}{4}\setcounter{bottomnumber}{3}\setcounter{totalnumber}{6}\setcounter{dbltopnumber}{3}
\renewcommand{\topfraction}{0.95}\renewcommand{\bottomfraction}{0.9}\renewcommand{\textfraction}{0.05}
\renewcommand{\floatpagefraction}{0.85}\renewcommand{\dbltopfraction}{0.95}\renewcommand{\dblfloatpagefraction}{0.85}
\section{Method Details}
\label{app:method}

\subsection{Formulas}
\label{app:formulas}

\paragraph{Expected accepted tokens.}
A node of a tree $\mathcal{T}$ is accepted when its path matches the tokens $x^\star$ that the target generates.
For a block of length $B$ with the candidates $C_i$ at position $i$, the expected number of accepted nodes is
\begin{equation}
\sum_{i=1}^{B-1} \sum_{c \in C_i} P\big(x^\star_i = c,\, x^\star_{1:i} \in \mathcal{T}\big),
\label{eq:yield}
\end{equation}
the sum over the nodes of the probability that the target accepts every token on the node's path \citep{li2024eagle2,wang2025opttree}.
Each node adds its own probability to this sum, so the $N{-}1$ nodes with the highest probabilities give the highest expected AL (\S\ref{sec:paths}).

\paragraph{Pair scores and path scores.}
Let $z_i$ and $h_i$ be the drafter's logits and hidden state at position $i$.
For a candidate $p$ at position $i{-}1$ and a candidate $c \in C_i$, DFlash2's path selector and \arbora{}'s path scorer compute the score
\begin{equation}
s_i(p,c) = z_i(c) + \sum_{j=1}^{r} U_{p,j}\,(W h_i)_j\,U'_{c,j},
\label{eq:selector}
\end{equation}
where $U, U' \in \mathbb{R}^{|V| \times r}$ are token embeddings over the vocabulary $V$ and $W$ projects the hidden state to rank $r$.
For $i=1$, $p$ is the anchor.
DFlash2 follows one path from the anchor $y_0=x_0$ and picks $y_i = \arg\max_{c\in C_i} s_i(y_{i-1},c)$ at each position.
The path scorer normalizes the scores over $C_i$ into the pair score
\begin{equation}
\pi_i(c \mid p) = \frac{\exp s_i(p,c)}{\sum_{c' \in C_i} \exp s_i(p,c')},
\label{eq:node}
\end{equation}
and the path score $S$ of Eq.~\ref{eq:pathscore} is the product of the pair scores along a path.

\paragraph{Loss of the Arbora experts.}
On each training block $b$, only the Arbora expert $e^*_b$ with the most accepted tokens is trained (\S\ref{sec:experts}).
The loss follows multiple choice learning (MCL) and is
\begin{equation}
L_{\mathrm{MCL}} = \frac{\sum_b \sum_{i=1}^{B-1} w_i\, \mathrm{CE}^{(e^*_b)}_{b,i}}{\sum_b \sum_{i=1}^{B-1} w_i},
\label{eq:mcl}
\end{equation}
where $\mathrm{CE}^{(e)}_{b,i}$ is the cross-entropy of Arbora expert $e$ at position $i$ of block $b$.
The weight $w_i = \exp(-(i-1)/\gamma)$ is larger at earlier positions, because a later token is accepted only after the earlier ones.
We use $\gamma=7$.

\subsection{Training the Path Scorer}
\label{app:treeloss}

\paragraph{The loss.}
Algorithm~\ref{alg:treeloss} computes the loss of \S\ref{sec:treeloss} on the candidates of one block, for $N$ tokens verified per step.
All targets use $\beta=1$ and $\lambda=0.1$.
Other values give nearly the same AL (Appendix~\ref{app:path-ablation}).
The loss is computed with $N=16$.
The path scorer is trained on the candidates of the shared drafter alone, so the candidate sets $C_i$ stay fixed.
Only $U$, $U'$, and $W$ of Eq.~\ref{eq:selector} receive gradients.
Their rank is $r=256$, as in the released DFlash2 path selector.

\begin{algorithm}[t]
\small
\caption{Tree loss on the candidates of one block.}
\label{alg:treeloss}
\begin{algorithmic}[1]
\Require candidates $C_1,\dots,C_{B-1}$; pair scores $\pi_i(c \mid p)$; prefix $x_{1:m}$; $N$, $\beta$, $\lambda$
\Statex \textit{Build the tree as at inference (choices carry no gradient)}
\State $\mathcal{P}_0 \gets \{x_0\}$, \ $S(x_0) \gets 1$, \ $\mathcal{A} \gets \emptyset$ \Comment{$x_0$: anchor}
\For{$i = 1,\dots,B{-}1$}
  \State $S(uc) \gets S(u)\,\pi_i(c \mid u_{i-1})$ for $u \in \mathcal{P}_{i-1}$, $c \in C_i$
  \State add these paths to $\mathcal{A}$; $\mathcal{P}_i \gets$ the $N{-}1$ best of them
\EndFor
\Statex \textit{Lift each target prefix above its threshold}
\State $L \gets 0$
\For{$k = 1,\dots,\min(m, N{-}1)$}
  \State $\mathcal{R}_k \gets$ paths in $\mathcal{A}$ that leave $x_{1:m}$ at position $\le k$
  \State $\tau_k \gets$ the $(N{-}k)$-th highest $S$ in $\mathcal{R}_k$; skip $k$ if none \Comment{threshold}
  \State $L \gets L - \log \sigma\big(\beta\,(\log S(x_{1:k}) - \log \tau_k)\big)$ \Comment{Eq.~\ref{eq:treeloss}}
\EndFor
\State \Return $L + \lambda\,L_{\mathrm{CE}}$
\end{algorithmic}
\end{algorithm}

\paragraph{The tree.}
While the tree is built as at inference, every scored path is added to a set $\mathcal{A}$.
The tree is the $N{-}1$ highest-scoring paths of $\mathcal{A}$.

\paragraph{The threshold.}
The threshold $\tau_k$ of \S\ref{sec:treeloss} is taken over the paths in $\mathcal{A}$.
A path that leaves the target's path after position $k$ extends $x_{1:k}$ and scores no higher.
The set $\mathcal{A}$ holds every path of the tree, so $\tau_k$ decides whether $x_{1:k}$ enters the tree.
When fewer than $N{-}k$ paths compete, there is no threshold and the prefix adds no loss.
Ties are broken as at inference.
The shorter path comes first, and siblings are ordered by candidate index.

\paragraph{Schedule.}
\lineadj{1}{0}
The path scorer starts from the trained path selector of its drafter.
On the six targets below, it is trained with AdamW at learning rate $10^{-4}$, weight decay 0, and gradient clipping at 1.0, with 60 sequences per step.
The learning rate rises over the first 2.5\% of the steps and then decays to zero with a cosine schedule.
On Qwen3-8B and Qwen3.8-27B, training runs for 4{,}000 steps on 240K of the target's responses.
On GLM-5.3-Flash, MiniMax-M3.1-Flash, Qwen3.6-35B-A3B, and Gemma-4-31B-it, it runs for 1{,}000 steps, one pass over 60K responses.
On Muse-Glimmer-30B, the path scorer comes from the comparison of learning rates below.

\paragraph{Comparison of learning rates.}
\lineadj{1}{0}
The comparison of learning rates in Table~\ref{tab:scorer-lr} uses a separate, shorter training of the path scorer.
The training data are the target's responses at $T=0$ to 20{,}000 prompts.
Each response has 128 tokens and is generated with thinking mode enabled.
The prompts are drawn at random from the training prompts, which exclude fixed-500.
Every response position is one training block, and 5\% of the requests are held out.
The path scorer starts from the released path selector and is trained with AdamW, weight decay 0.01, 2{,}048 blocks per step, and 150 linear warm-up steps.
Each request is used once.
We train at the learning rates $10^{-5}$, $3\times10^{-5}$, $10^{-4}$, and $3\times10^{-4}$.
For each learning rate, the step is chosen by the AL on the held-out requests.
The path scorer of Muse-Glimmer-30B in Tables~\ref{tab:selector} and~\ref{tab:more-targets} comes from this comparison, at learning rate $10^{-5}$ and step 1{,}100, which give the highest AL on the held-out requests.

\subsection{Training Details}
\label{app:training}

\paragraph{Data and optimizer.}
All training uses the target's own responses at $T=0$.
We use 800K responses on Qwen3-8B and Qwen3.8-27B and 60K on each of GLM-5.3-Flash, MiniMax-M3.1-Flash, Qwen3.6-35B-A3B, and Gemma-4-31B-it.
On these four targets, the 60K responses answer the same prompts, which are among the training prompts of Qwen3.8-27B.
Those of GLM-5.3-Flash and MiniMax-M3.1-Flash have up to 4{,}096 tokens.
Generating them takes about 4 hours on 16 GPUs for GLM-5.3-Flash, and on eight GPUs it takes 5.4 hours for MiniMax-M3.1-Flash, 3.6 hours for Qwen3.6-35B-A3B, and 4.5 hours for Gemma-4-31B-it.
Sequences are cut at 3{,}072 tokens on Qwen3-8B and at about 8K on the other targets.
Each training step uses 60 sequences and up to 512 blocks of 16 tokens per sequence, each block starting at a randomly chosen anchor.
The drafters and the Arbora experts are also trained with AdamW, weight decay 0, and gradient clipping at 1.0.

\paragraph{Drafters.}
The Qwen3-8B DFlash2 drafter starts from the released Qwen3-8B DFlash drafter, which we first train for five epochs on the 800K responses.
We add DFlash2's convolutions, initialized to the identity, and its path selector, initialized at random.
We then train the whole drafter with DFlash2's loss for one more epoch of 13{,}325 steps.
The learning rate rises to $10^{-3}$ over the first 0.5\% of the steps and then decays to zero with a cosine schedule.
The Qwen3.8-27B drafter is the released DFlash2 drafter trained for three more epochs at learning rate $10^{-4}$.
Both are trained with blocks of 16.
With these two drafters, DFlash2 has a higher AL on fixed-500 than with the released DFlash2 drafters of the same targets (Tables~\ref{tab:fixed500} and~\ref{tab:released}).
On the other targets, we use the released DFlash2 drafters without training them.

\paragraph{Arbora experts.}
On these six targets, $E=8$ Arbora experts are trained with the loss of Eq.~\ref{eq:mcl}, while the shared drafter and its path selector stay frozen.
The hidden dimension of an Arbora expert is 1{,}536 on Qwen3-8B, GLM-5.3-Flash, and MiniMax-M3.1-Flash, 2{,}176 on Qwen3.8-27B, 768 on Qwen3.6-35B-A3B, and 2{,}688 on Gemma-4-31B-it.
Each Arbora expert has a gate, an up, and a down projection.
The gate and up projections are initialized from $\mathcal{N}(0, 0.02^2)$ and the down projection from $\mathcal{N}(0, 0.001^2)$.
When several Arbora experts have the most accepted tokens, the one with the lowest weighted cross-entropy is trained.
The weights are those of Eq.~\ref{eq:mcl}.
On Qwen3-8B, the Arbora experts are trained for 2{,}000 steps on 120K sequences.
The learning rate is $10^{-3}$ after 30 warm-up steps, stays constant until step 1{,}600, and then decays along a cosine curve to half of this value at step 2{,}000.
On the other five targets, they are trained for 1{,}000 steps on 60K sequences.
The learning rate is $6\times10^{-4}$ after 11 warm-up steps, stays constant until step 800, and then decays along a cosine curve to half of this value at step 1{,}000.

\paragraph{What each method trains.}
DFlash2 trains the whole drafter and its path selector.
\arbora{} keeps this drafter frozen and trains only the path scorer and the Arbora experts, so the two methods share every weight of the drafter.
The gain of \arbora{} does not come from the additional training steps (Appendix~\ref{app:experts-design}).

\paragraph{Checkpoints.}
On these six targets, every path scorer and every set of Arbora experts is taken at its last training step.
No choice is made on the public benchmarks or on the measurements with released drafters.

\paragraph{How the Arbora experts differ from each other.}
\lineadj{-1}{0}
Table~\ref{tab:wins} shows how often each Arbora expert is the best one on a training block.
Each is the best on 10--17\% of the blocks, against 12.5\% for an even split, without a router or a balance term.
All eight therefore keep receiving updates.
The Arbora experts differ in which of the likely tokens they rank first.
On Qwen3-8B, fixed-500, one Arbora expert alone has the most accepted tokens in 4{,}471 of 10{,}493 blocks.
In these blocks, take the first position where the draft of the second-best Arbora expert is not correct.
This Arbora expert ranks the correct token second in 48.6\% of the blocks, third in 18.7\%, and in its top 8 in 91.3\%.

\paragraph{Where the Arbora experts differ from the shared drafter.}
Table~\ref{tab:first-token} groups the positions by the ratio $p_2/p_1$ of the probabilities of the shared drafter's second and top tokens.
The Arbora experts' top tokens differ from the shared drafter's mostly in the rows with a high ratio, where the shared drafter is unsure.

\begin{table}[ht]
\centering
\small
\setlength{\tabcolsep}{2.5pt}
\resizebox{\columnwidth}{!}{%
\begin{tabular}{@{}l r r r r r r@{}}
\toprule
& & \multicolumn{3}{c}{Fully trained} & \multicolumn{2}{c}{Early step} \\
\cmidrule(lr){3-5}\cmidrule(l){6-7}
$p_2/p_1$ & Positions & Per expert & Any & Top & Best & Random \\
\midrule
0--0.01 & 11.9\% & 2.9\% & 17.3\% & 1.20 & 2.4\% & 0.0\% \\
0.01--0.05 & 6.2\% & 18.2\% & 78.3\% & 2.15 & 15.3\% & 0.0\% \\
0.05--0.1 & 4.0\% & 27.7\% & 91.0\% & 2.63 & 23.9\% & 0.0\% \\
0.1--0.25 & 9.5\% & 35.0\% & 92.7\% & 2.95 & 31.2\% & 0.0\% \\
0.25--0.5 & 19.1\% & 41.3\% & 90.4\% & 3.21 & 38.4\% & 0.0\% \\
0.5--0.75 & 20.8\% & 47.1\% & 92.5\% & 3.42 & 45.2\% & 0.0\% \\
0.75--1 & 28.5\% & 59.9\% & 98.7\% & 3.76 & 58.8\% & 8.6\% \\
\bottomrule
\end{tabular}}
\caption{Arbora experts' top tokens compared with the shared drafter's on Qwen3-8B, fixed-500, over 892{,}500 positions, each draft with its own probabilities. Rows group the positions by the ratio $p_2/p_1$ of the probabilities of the shared drafter's second and top tokens. Positions: share of positions in each row. Per expert: how often an Arbora expert's top token differs from the shared drafter's. Any: how often at least one of the eight differs. Top: mean number of distinct top tokens among the eight. These three columns use the fully trained Arbora experts. Best and Random: the per-expert rate at an early training step, when each block trains the best Arbora expert or a randomly chosen one.}
\label{tab:first-token}
\end{table}

\begin{table}[ht]
\centering
\small
\setlength{\tabcolsep}{2.6pt}
\begin{tabular}{@{}l r r r r r r r r@{}}
\toprule
Arbora expert & 1 & 2 & 3 & 4 & 5 & 6 & 7 & 8 \\
\midrule
Qwen3-8B & 14.1 & 13.7 & 13.0 & 14.3 & 13.2 & 10.6 & 10.2 & 10.9 \\
Qwen3.8-27B & 10.6 & 15.6 & 16.7 & 13.1 & 10.2 & 11.4 & 10.3 & 12.0 \\
\bottomrule
\end{tabular}
\caption{Share of training blocks (\%) on which each Arbora expert is the best of the eight, over the 2{,}000 training steps on Qwen3-8B and the 1{,}000 on Qwen3.8-27B.}
\label{tab:wins}
\end{table}

\subsection{Merging the Arbora Experts' Paths}
\label{app:merge}

\begin{table}[ht]
\centering
\small
\begin{tabular}{@{}r r r r@{}}
\toprule
$N$ & sum (ours) & maximum & union rescored \\
\midrule
16 & \textbf{5.671} & 5.559 & 5.659 \\
64 & \textbf{6.844} & 6.789 & 6.838 \\
\bottomrule
\end{tabular}
\caption{Trace-driven AL of the eight Qwen3-8B Arbora experts on fixed-500, averaged over its five domains, under three merge rules. In all three, each Arbora expert first lists its own $N{-}1$ best paths from its top 8 tokens per position. Sum: the score of a path is the sum of the probabilities that it has under the Arbora experts that listed it. Maximum: the score is the largest of these probabilities. Union rescored: every listed path is rescored by all eight Arbora experts, with tokens outside an Arbora expert's top 64 counted as zero.}
\label{tab:merge}
\end{table}

\lineadj{1}{0}
Table~\ref{tab:merge} compares three rules for merging the paths.
\arbora{} uses the sum rule.
It has the highest AL at $N=16$ and at $N=64$, and it needs no rescoring of the listed paths.
An Arbora expert that lists a path also lists all its prefixes, each with at least the path's score.
Under the sum rule, a prefix therefore scores at least as high as its extensions, and the $N{-}1$ highest-scoring merged paths form a tree.
The sum rule gives a path no score from an Arbora expert that ranks it just outside its own $N{-}1$ best paths.
Rescoring the union adds these probabilities and can admit such a path.
The admitted nodes are 2.5\% and 2.6\% of the rescored trees at $N=16$ and 64.
They are accepted no more often than the nodes they replace, and the AL is 0.1--0.2\% lower than under the sum rule.

\section{Evaluation Setup}
\label{app:setup}

\subsection{Parameter Settings}
\label{app:rounds}

\paragraph{Decoding.}
Each fixed-500 request generates up to 128 tokens, and each public-benchmark request up to 512.
At $T=1$, the target samples without top-p or top-k, and every method drafts as at $T=0$.
With one request at a time, throughput (tok/s) is the number of output tokens divided by the sum of the requests' decoding times.

\paragraph{Drafters.}
The drafter gives $K=16$ candidates per position.
On MiniMax-M3.1-Flash it gives 32, the setting of its released drafter.
The released DFlash2 drafters of Qwen3-8B, Qwen3.8-27B, Qwen3.6-35B-A3B, Gemma-4-31B-it, GLM-5.3-Flash, and MiniMax-M3.1-Flash were trained with blocks of 8, and we run them with blocks of 16.
Running with blocks of 16 raises the AL of DFlash2 with the released Qwen3.8-27B drafter from 4.43 to 4.84.
The Qwen3.8-27B drafter of Table~\ref{tab:main} is the released drafter trained further with blocks of 16.
The Qwen3-8B drafter of Table~\ref{tab:main} is our own DFlash2 drafter, also trained with blocks of 16 (Appendix~\ref{app:training}).
On every target, DFlash2 and \arbora{} run on the same drafter.

\paragraph{Implementation.}
Qwen3-8B runs on one H100.
GLM-5.3-Flash and MiniMax-M3.1-Flash run on eight and the other targets on two, with tensor parallelism.
All methods use the same SGLang implementation and the same precision for the target, the drafter, and the path selector or path scorer.
We optimize only the parts that \arbora{} adds, which are tree construction, the tree mask, acceptance along the tree, the path scoring of tree nodes, and the Arbora experts.
The Arbora experts are stored in INT4.
The Arbora experts run on a second CUDA stream beside the frozen MLP, which runs on cuBLAS.
They use our own INT4 kernels on Qwen3-8B, Qwen3.8-27B, Qwen3.6-35B-A3B, and Gemma-4-31B-it for up to 128 rows, and Marlin \citep{frantar2025marlin} beyond 128 rows and on GLM-5.3-Flash and MiniMax-M3.1-Flash.
For targets with Gated DeltaNet layers \citep{yang2025gated}, each tree node starts from the state of its parent, and the state of the last accepted node is kept.

\paragraph{Rounds.}
In a round, the methods that are compared run one after another on the same GPUs.
A pass serves every prompt of the workload once.
Every run starts a new server and sends ten warm-up requests that are not timed.
The GPUs are in a shared cluster, where jobs of other users can slow a run down but never speed it up.
The fastest rounds are therefore the least disturbed, and we report throughput from them.
Every method is measured with the same protocol, and the methods that are compared alternate within each round.
A change in the load of the cluster therefore affects them alike, and the comparison between them is fair.
On fixed-500, averaging all passes gives the same gains, as the paragraph on the variation across rounds shows.

\paragraph{fixed-500 at $T=0$.}
Every method makes four passes over fixed-500, two in each of two rounds.
DFlash2, \arboraOne{}, and \arbora{} run one after another on the same GPUs in each round.
The two path scorers of Table~\ref{tab:selector}, before and after training, run one after another in two rounds of their own.
The throughput is the mean of the three fastest of the four passes, and the AL is the mean of the four.
Table~\ref{tab:domains} splits the passes of DFlash2 and \arbora{} by domain.

\paragraph{fixed-500 at $T=1$.}
Every method again makes four passes, one in each of four rounds, and round $k$ uses random seed $k$.
The throughput is the mean of the three fastest of the four passes, and the AL is the mean of the four.

\begin{table*}[t]
\centering
\footnotesize
\setlength{\tabcolsep}{2.2pt}
\begin{tabular}{@{}l l l c r r r r r r r r@{}}
\toprule
& & & & \multicolumn{4}{c}{$T=0$} & \multicolumn{4}{c}{$T=1$} \\
\cmidrule(lr){5-8}\cmidrule(l){9-12}
Target & Method & Scores & Experts & AL & $\Delta$ & tok/s & $\Delta$ & AL & $\Delta$ & tok/s & $\Delta$ \\
\midrule
\multirow{4}{*}{Qwen3.8-27B} & DFlash2 & path selector & 0 & 5.03 & -- & 269.0 & -- & 3.79 & -- & 206.7 & -- \\ %
& \arboraOne{} & path scorer & 0 & 6.26 & +24.5\% & 328.9 & +22.3\% & 4.79 & +26.5\% & \textbf{259.0} & \textbf{+25.3\%} \\ %
& \arboraEp{} & per position & 8 & 6.05 & +20.3\% & 313.6 & +16.6\% & 4.69 & +24.0\% & 249.3 & +20.6\% \\ %
& \textbf{\arbora{}} & path scorer & 8 & \textbf{6.40} & \textbf{+27.2\%} & \textbf{329.3} & \textbf{+22.4\%} & \textbf{4.81} & \textbf{+27.0\%} & 254.6 & +23.2\% \\ %
\midrule
\multirow{4}{*}{Qwen3-8B} & DFlash2 & path selector & 0 & 4.76 & -- & 514.2 & -- & 4.31 & -- & 464.1 & -- \\ %
& \arboraOne{} & path scorer & 0 & 5.95 & +25.1\% & \textbf{624.3} & \textbf{+21.4\%} & 5.35 & +24.3\% & \textbf{561.7} & \textbf{+21.0\%} \\ %
& \arboraEp{} & per position & 8 & 5.61 & +17.9\% & 567.8 & +10.4\% & 5.15 & +19.6\% & 520.3 & +12.1\% \\ %
& \textbf{\arbora{}} & path scorer & 8 & \textbf{5.98} & \textbf{+25.7\%} & 609.2 & +18.5\% & \textbf{5.46} & \textbf{+26.8\%} & 555.9 & +19.8\% \\ %
\midrule
\multirow{4}{*}{GLM-5.3-Flash} & DFlash2 & path selector & 0 & 4.43 & -- & 369.5 & -- & 3.98 & -- & 332.8 & -- \\ %
& \arboraOne{} & path scorer & 0 & 5.25 & +18.5\% & 439.7 & +19.0\% & 4.71 & +18.2\% & 396.4 & +19.1\% \\ %
& \arboraEp{} & per position & 8 & 5.52 & +24.6\% & 449.5 & +21.7\% & 5.00 & +25.5\% & 408.0 & +22.6\% \\ %
& \textbf{\arbora{}} & path scorer & 8 & \textbf{5.68} & \textbf{+28.2\%} & \textbf{463.8} & \textbf{+25.5\%} & \textbf{5.12} & \textbf{+28.5\%} & \textbf{418.2} & \textbf{+25.7\%} \\ %
\midrule
\multirow{4}{*}{\begin{tabular}[c]{@{}l@{}}MiniMax-\\M3.1-Flash\end{tabular}} & DFlash2 & path selector & 0 & 3.59 & -- & 338.3 & -- & 3.14 & -- & 295.8 & -- \\ %
& \arboraOne{} & path scorer & 0 & 4.52 & +25.8\% & \textbf{419.2} & \textbf{+23.9\%} & 3.96 & +26.3\% & 369.3 & +24.8\% \\ %
& \arboraEp{} & per position & 8 & 4.58 & +27.6\% & 403.5 & +19.3\% & 4.06 & +29.6\% & 359.3 & +21.4\% \\ %
& \textbf{\arbora{}} & path scorer & 8 & \textbf{4.74} & \textbf{+31.9\%} & 418.3 & +23.7\% & \textbf{4.18} & \textbf{+33.3\%} & \textbf{370.9} & \textbf{+25.4\%} \\ %
\bottomrule
\end{tabular}
\caption{Ablation on the four main targets. One request, fixed-500, 16 tokens verified per step. $\Delta$ is the change relative to DFlash2, computed from unrounded values. The Scores column says whether the candidates are scored by DFlash2's path selector, by each draft's own probabilities at each position (per position), or by \arbora{}'s path scorer. The Experts column gives the number of Arbora experts.}
\label{tab:ablation}
\end{table*}

\paragraph{Public benchmarks.}
DFlash2 and \arbora{} run one after another in three rounds, and their order changes in every round.
At $T=0$, every cell reports its fastest round, with the AL of that round.
At $T=0$ the AL differs little between rounds, with a standard deviation of at most 0.024 on fixed-500.
At $T=1$, round $k$ uses random seed $k$, and every cell reports the throughput of its fastest round and the mean AL of the three rounds.
The target alone runs one round at $T=0$, and Tables~\ref{tab:public} and~\ref{tab:public-t1} both use this round.
On Qwen3-8B and Qwen3.8-27B, the benchmarks are split into groups that run at the same time on separate GPUs of one node, and each benchmark has its own servers.
On GLM-5.3-Flash and MiniMax-M3.1-Flash, they run one after another.

\paragraph{Other measurements.}
The measurements under load, with released drafters, and over the number of tokens verified per step alternate between methods in rounds.
They run five rounds under load and three otherwise.
Every cell reports its fastest round.
A cell with concurrent requests serves fixed-500 three times, 1{,}500 requests.
On Qwen3.8-27B, the scans over tokens verified per step measure the AL.
On this target, the AL at $T=0$ is the same in every round, so they run one round.

\paragraph{Variation across rounds.}
\lineadj{1}{0}
Over the four passes of a table row on fixed-500 at $T=0$, the standard deviation of the throughput is at most 0.5\% of its mean, and that of the AL is at most 0.024.
Over the four rounds at $T=1$, the standard deviation of the AL is at most 0.05.
The mean throughput gain of \arbora{} over DFlash2 on fixed-500 hardly changes with how the passes are summarized.
At $T=0$, the mean of all four passes gives 22.5\% and each method's median throughput gives 22.5\%, against the reported 22.5\%.
At $T=1$, they give 23.5\% and 23.6\%, against 23.5\%.

\subsection{Trace-Driven Evaluation}
\label{app:offline}

\paragraph{Traces.}
The target decodes fixed-500 alone at $T=0$.
Every output position is the anchor of one step.
At each of the 15 drafted positions of its block, we record the drafter's candidates with the scores of the DFlash2 path selector and of the path scorer.
The target's output does not depend on the drafter, so every method can be evaluated on the same outputs without serving.

\paragraph{Replay.}
\lineadj{1}{0}
Each method then decodes as in serving.
At the current anchor, it builds its path or tree of $N$ tokens from the recorded candidates and its own scores, where $N$ is the number of tokens verified per step.
It accepts the longest prefix that matches the target's tokens, appends the bonus token, and moves the anchor past them.
The trace-driven AL is the number of output tokens divided by the number of steps.
The replay uses the drafter, the path scorer, and the Arbora experts with which the traces were recorded.
These weights are not those of Tables~\ref{tab:ablation} and~\ref{tab:fixed500}, so trace-driven AL values are compared only with each other.
On Qwen3.8-27B at $N=16$, the trace-driven AL is within 0.03 of the serving AL of the same weights.
On Qwen3-8B, it is about 2\% higher for both \arboraOne{} and \arbora{}.

\paragraph{Where acceptance stops.}
\lineadj{0}{1}
A step ends in one of three cases.
In the first, the correct token at the rejected position is among the candidates.
In the second, it is not among them.
In the third, the whole block is accepted or the request ends.
With $N=16$ on Qwen3.8-27B, the shares of the three cases are 77.2\%, 12.4\%, and 10.4\% for DFlash2.
The two shares in \S\ref{sec:background} are the first two cases of DFlash2, taken as fractions of their sum.
For the shares of \S\ref{sec:background} with 4, 8, or 16 candidates, only the first 4, 8, or 16 of the 16 candidates recorded per position on Qwen3.8-27B are kept.
The correct token is among them in 58\%, 75\%, and 86\% of the steps of DFlash2 that stop before the end of the block, and the AL is 5.03, 5.07, and 5.07.
Verifying the target's own tokens for as long as they are among the candidates gives an upper bound on the AL of any scoring.
On Qwen3.8-27B, this bound is 9.14 for every $N$ from 16 on, against 5.07 for DFlash2 at $N=16$.

\section{Ablation}
\label{app:ablations}

\subsection{Per Target}
\label{app:ablation}

Table~\ref{tab:ablation} compares DFlash2, the path scorer alone, the Arbora experts alone, and both together on the four main targets (\S\ref{sec:results-analysis}).
On every target and at both $T=0$ and $T=1$, the path scorer and the Arbora experts each raise the AL over DFlash2 on their own, and both together give the highest AL.
The AL gain of the Arbora experts over the path scorer alone grows with the number of tokens verified per step (Appendix~\ref{app:budget}).
Appendix~\ref{app:experts-design} compares the Arbora experts with the shared drafter alone under the same scores.

\subsection{Path Scorer}
\label{app:path-ablation}

Table~\ref{tab:selector} gives \arboraOne{} on each target before and after the path scorer is trained with the tree loss on the shared drafter's trees (Appendix~\ref{app:treeloss}).
The training raises the AL and the throughput on every target.
On the four main targets, \arboraOne{} is 18.7\% faster than DFlash2 on average before the training and 21.6\% after it, with the DFlash2 throughput of Table~\ref{tab:ablation}.

\begin{table}[ht]
\centering
\small
\setlength{\tabcolsep}{4pt}
\begin{tabular}{@{}l r r r r@{}}
\toprule
& \multicolumn{2}{c}{Before} & \multicolumn{2}{c}{After} \\
\cmidrule(lr){2-3}\cmidrule(l){4-5}
Target & AL & tok/s & AL & tok/s \\
\midrule
Qwen3-8B & 5.834 & 612.2 & 5.954 & 623.7 \\
Qwen3.8-27B & 6.135 & 323.5 & 6.264 & 328.8 \\
GLM-5.3-Flash & 5.073 & 424.0 & 5.252 & 439.5 \\
MiniMax-M3.1-Flash & 4.399 & 408.8 & 4.523 & 419.8 \\
Qwen3.6-35B-A3B & 6.858 & 705.7 & 6.991 & 718.8 \\
Muse-Glimmer-30B & 8.352 & 478.3 & 8.416 & 482.7 \\
Gemma-4-31B-it & 6.215 & 260.4 & 6.282 & 262.3 \\
\bottomrule
\end{tabular}
\caption{\arboraOne{} before and after the path scorer is trained with the tree loss, on the same shared drafter. $T=0$, one request, fixed-500, 16 tokens verified per step. Before, the path scorer has the weights of DFlash2's path selector for this drafter, trained with cross-entropy given the correct previous token.}
\label{tab:selector}
\end{table}

\paragraph{The objective.}
In Table~\ref{tab:objective}, the same path selector is trained once with DFlash2's cross-entropy and once with the tree loss at $N=16$ tokens verified per step, on the same data and for the same number of steps.
The tree loss gives the highest AL at every $N$, so the gain comes from the objective.
On Qwen3-8B, the tree loss gives a 1.0\% higher AL than cross-entropy at $N=16$, 32, and 64.

\begin{table}[ht]
\centering
\small
\setlength{\tabcolsep}{4pt}
\begin{tabular}{@{}l r r r r@{}}
\toprule
Path selector & $N=8$ & $N=16$ & $N=32$ & $N=64$ \\
\midrule
Before training & 4.817 & 6.142 & 6.823 & 7.313 \\
Cross-entropy & 4.825 & 6.139 & 6.819 & 7.298 \\
Tree loss & \textbf{4.859} & \textbf{6.213} & \textbf{6.900} & \textbf{7.392} \\
\bottomrule
\end{tabular}
\caption{Trace-driven AL of \arboraOne{} on Qwen3.8-27B, fixed-500, with the path selector before training and after the same number of training steps with each loss.}
\label{tab:objective}
\end{table}

\paragraph{The loss settings.}
\lineadj{1}{0}
On Qwen3.8-27B, we vary the scaling constant $\beta$ of the tree loss and the weight $\lambda$ of its cross-entropy term, and train each setting from the same start on the same data.
We change one of them at a time, to $\beta=0.5$ or 2 and to $\lambda=0$ or 1, and we also train the tree loss at $N=8$ to 64 together.
The trace-driven AL at $N=16$ is 6.19 or 6.20 in every setting, including the $\beta=1$ and $\lambda=0.1$ that we use.
Cross-entropy gives 6.13, and the path selector before training gives 6.14.

\paragraph{The scores and the selection.}
\lineadj{1}{0}
In Table~\ref{tab:scorer-structure}, one part of the path scorer is changed at a time on Qwen3.8-27B.
The single path takes the path scorer's best next candidate at each position, as DFlash2 does with its path selector.
Taking each candidate's previous token from the single path instead of from its own path lowers the AL by 5.4\% at $N=16$ and by 13.2\% at $N=128$, so the pair scores must follow the path.
Always keeping the single path in the tree lowers the AL by up to 17.1\%, so the selection must be free to leave it.
Halving the number $K$ of candidates per position lowers the AL by up to 4.3\%.
Doubling it adds at most 1.5\%, less than the Arbora experts add at the same $N$ (Table~\ref{tab:nscan}).

\begin{table}[ht]
\centering
\small
\setlength{\tabcolsep}{2pt}
\begin{tabular}{@{}l r r r r r@{}}
\toprule
& \multicolumn{5}{c}{$N$} \\
\cmidrule(l){2-6}
Variant & 8 & 16 & 32 & 64 & 128 \\
\midrule
Path scorer, $K=16$ & 4.860 & 6.211 & 6.902 & 7.392 & 7.756 \\
Previous from single path & 4.776 & 5.876 & 6.315 & 6.579 & 6.731 \\
Single path kept in tree & 4.512 & 5.150 & 6.724 & 7.339 & 7.738 \\
$K=8$ & 4.844 & 6.172 & 6.785 & 7.154 & 7.424 \\
$K=32$ & 4.862 & 6.225 & 6.912 & 7.445 & 7.869 \\
\bottomrule
\end{tabular}
\caption{Trace-driven AL of \arboraOne{} on Qwen3.8-27B, fixed-500. The path scorer is not retrained for the variants.}
\label{tab:scorer-structure}
\end{table}

\paragraph{The learning rate.}
Table~\ref{tab:scorer-lr} gives the trace-driven AL of the path scorer trained at each learning rate on five targets (Appendix~\ref{app:treeloss}).
At learning rates $10^{-5}$ and $3\times10^{-5}$, the path scorer has a higher AL than the path selector of the released drafter on each of the five targets.

\begin{table}[ht]
\centering
\small
\setlength{\tabcolsep}{3pt}
\resizebox{\columnwidth}{!}{%
\begin{tabular}{@{}l r r r r r@{}}
\toprule
Target & Released & $10^{-5}$ & $3\times10^{-5}$ & $10^{-4}$ & $3\times10^{-4}$ \\
\midrule
GLM-5.3-Flash & 5.075 & 5.193 & \textbf{5.233} & 5.228 & 5.143 \\
MiniMax-M3.1-Flash & 4.414 & 4.499 & \textbf{4.517} & 4.467 & 4.456 \\
Qwen3.6-35B-A3B & 6.871 & 6.968 & \textbf{6.994} & 6.946 & 6.881 \\
Muse-Glimmer-30B & 8.321 & \textbf{8.400} & 8.381 & 8.309 & 8.272 \\
Gemma-4-31B-it & 6.196 & 6.201 & \textbf{6.217} & 6.206 & 6.169 \\
\bottomrule
\end{tabular}}
\caption{Trace-driven AL on fixed-500 at $N=16$ with the path scorer trained at each learning rate. Released is the path selector of the released drafter. Bold marks the highest AL in each row.}
\label{tab:scorer-lr}
\end{table}

\subsection{Arbora Experts}
\label{app:experts-design}

Table~\ref{tab:experts} compares ways of adding parameters to the shared drafter on Qwen3-8B, at $N=16$, 32, and 64 tokens verified per step.

\begin{table}[ht]
\centering
\small
\setlength{\tabcolsep}{4pt}
\begin{tabular}{@{}l r r r@{}}
\toprule
Drafter (Qwen3-8B) & $N=16$ & $N=32$ & $N=64$ \\
\midrule
Shared drafter only & 5.147 & 5.640 & 5.997 \\
+ one MLP of width 12{,}288 & 5.147 & 5.635 & 6.013 \\
\multicolumn{4}{@{}l}{+ Arbora experts} \\
\quad 4, width 1/8 & 5.512 & 6.111 & 6.581 \\
\quad 4, width 1/4 & 5.505 & 6.117 & 6.569 \\
\quad 8, width 1/8 & 5.598 & 6.259 & 6.725 \\
\quad 8, width 1/8, random & 5.148 & 5.632 & 6.008 \\
\quad 8, width 1/8, fully trained & \textbf{5.670} & \textbf{6.338} & \textbf{6.843} \\
\midrule
Drafter, trained further & 5.161 & 5.619 & 5.959 \\
5 separately trained drafters & 5.235 & 5.739 & 6.138 \\
\bottomrule
\end{tabular}
\caption{Trace-driven AL on fixed-500, averaged over its five domains. Each drafter is scored with its own probabilities. Added parameters are trained beside the shared drafter's frozen MLP. The wide MLP and the last row of Arbora experts are fully trained, and the other rows of Arbora experts stop at the same early step. Each row of Arbora experts merges one draft per Arbora expert, with no draft of the shared drafter alone. Random: as the row above, except that each block trains a randomly chosen Arbora expert instead of the best one. The last row merges five separately trained drafters into one tree with SpecInfer's merge.}
\label{tab:experts}
\end{table}

\paragraph{Added parameters.}
\lineadj{0}{1}
Eight fully trained Arbora experts raise the AL over the shared drafter alone by 10.2\% at $N=16$.
One wide MLP with the same number of parameters leaves the AL unchanged.
Training the drafter itself for about as many additional steps as the Arbora experts are trained changes its AL by less than 0.7\%, so the gain does not come from more training.
Five drafters trained separately gain 1.7\% at $N=16$ when SpecInfer's merge combines them into one tree.
They need five drafter passes per step.
The attention and MLP of one drafter pass take 0.95\,ms of an 8.62\,ms step (Table~\ref{tab:latency}), so the four additional passes would lengthen the step by about 44\% for a 1.7\% higher AL.

\paragraph{Number and width.}
In our tests, the number of Arbora experts matters, and their width does not.
With the same number of parameters, eight Arbora experts of width 1/8 have a 1.7--2.4\% higher AL than four of width 1/4.
Doubling the width of four Arbora experts leaves the AL unchanged.

\paragraph{Choosing the best Arbora expert.}
The gain depends on training only the best Arbora expert on each block.
Training a randomly chosen Arbora expert on each block instead, with everything else unchanged, gives the same AL as the shared drafter alone (Table~\ref{tab:experts}, the random row).
The top tokens of Arbora experts trained in this way rarely differ from the shared drafter's (Table~\ref{tab:first-token}, the Random column).

\paragraph{The training loss.}
Training the Arbora experts with the tree loss on their merged tree, instead of with the best-expert loss of Eq.~\ref{eq:mcl}, changes the AL at the same step by $-$0.2\% to +0.7\% and takes 3.3 times as long.

\begin{table}[ht]
\centering
\small
\setlength{\tabcolsep}{3pt}
\begin{tabular}{@{}r r r r@{}}
\toprule
$N$ & Top 8 & Top 16 & Top 32 \\
\midrule
\multicolumn{4}{@{}l}{\textit{Qwen3-8B}} \\
16 & 104\% & 105\% & -- \\
32 & 88\% & 99\% & -- \\
64 & 62\% & 94\% & -- \\
\midrule
\multicolumn{4}{@{}l}{\textit{Qwen3.8-27B}} \\
16 & 80\% & 105\% & -- \\
32 & 39\% & 87\% & -- \\
64 & $-$2\% & 76\% & -- \\
\midrule
\multicolumn{4}{@{}l}{\textit{MiniMax-M3.1-Flash}} \\
16 & 56\% & 79\% & 88\% \\
32 & 36\% & 73\% & 90\% \\
64 & 2\% & 48\% & 78\% \\
\bottomrule
\end{tabular}
\caption{Share of the Arbora experts' trace-driven AL gain over the shared drafter that remains when the Arbora experts are restricted at each position to the shared drafter's top 8, 16, or 32 tokens, with the path scorer. Each draft gives 16 candidates on Qwen3-8B and Qwen3.8-27B and 32 on MiniMax-M3.1-Flash.}
\label{tab:experts-topk}
\end{table}

\paragraph{Where the Arbora experts' extra candidates lie.}
\lineadj{1}{1}
On Qwen3-8B, with each draft scored by its own probabilities, we take every output position of fixed-500 as an anchor.
The Arbora experts' tree holds 13.16 distinct candidates at $N=16$, against 11.22 in the shared drafter's tree, and 32.92 against 24.68 at $N=64$.
Of the Arbora experts' candidates, 0.34 at $N=16$ and 4.27 at $N=64$ lie outside the shared drafter's top 8 tokens.
We group the anchors by the correct token that the shared drafter's tree does not hold when it stops.
Where this token is in the shared drafter's top 8, the Arbora experts' tree accepts 5.62 tokens against the shared drafter's 4.88 at $N=16$, and 7.25 against 5.83 at $N=64$.
These anchors are 77.1\% of all anchors at $N=16$ and 64.1\% at $N=64$, and they carry all of the net gain in accepted tokens at $N=16$ and 93\% at $N=64$.

\lineadj{1}{0}
With each draft scored by its own probabilities, restricting each Arbora expert to the shared drafter's top 8 tokens keeps 98\%, 95\%, and 90\% of the fully trained Arbora experts' AL gain over the shared drafter at $N=16$, 32, and 64.
Restricting it to the shared drafter's top 64 tokens changes the AL by at most 0.1\%.
With the path scorer that \arbora{} uses, the share of the gain that needs tokens outside the shared drafter's top 8 grows with $N$ and differs between targets (Table~\ref{tab:experts-topk}).

\paragraph{Merging.}
\lineadj{1}{0}
The Arbora experts raise the AL only when the merge combines their whole paths.
Averaging the Arbora experts' probabilities at each position and building one tree from them gives an AL of 4.90 at $N=16$ and 5.81 at $N=64$, lower than the shared drafter's 5.147 and 5.997.

\subsection{Tokens Verified per Step}
\label{app:budget}

\begin{table}[!ht]
\centering
\small
\setlength{\tabcolsep}{3.5pt}
\begin{tabular}{@{}l r r r@{}}
\toprule
& \multicolumn{2}{c}{Qwen3-8B} & Qwen3.8-27B \\
\cmidrule(lr){2-3}
Method, $N=64$ & AL & tok/s & AL \\
\midrule
\arboraOne{} & 7.29 & 697.9 & 7.48 \\
\arboraEp{} & 6.68 & 624.4 & 7.13 \\
\arboraE{} & \textbf{7.53} & \textbf{699.8} & \textbf{7.89} \\
\midrule
DFlash2 ($N=16$) & 4.76 & 514.2 & 5.03 \\
\bottomrule
\end{tabular}
\caption{AL and throughput (tok/s) at $T=0$ with 64 tokens verified per step, one request, fixed-500, settings of Appendix~\ref{app:rounds}.}
\label{tab:budget}
\end{table}

\begin{table}[!ht]
\centering
\small
\setlength{\tabcolsep}{3pt}
\begin{tabular}{@{}r r r r r@{}}
\toprule
& \multicolumn{2}{c}{Qwen3-8B} & \multicolumn{2}{c}{Qwen3.8-27B} \\
\cmidrule(lr){2-3}\cmidrule(l){4-5}
$N$ & $E{=}0$ & $E{=}8$ & $E{=}0$ & $E{=}8$ \\
\midrule
16 & 5.95 & 5.98 {\scriptsize +0.4\%} & 6.26 & 6.40 {\scriptsize +2.2\%} \\
32 & 6.68 & 6.87 {\scriptsize +2.7\%} & 6.98 & 7.27 {\scriptsize +4.1\%} \\
64 & 7.29 & 7.53 {\scriptsize +3.2\%} & 7.48 & 7.89 {\scriptsize +5.4\%} \\
128 & 7.73 & 8.11 {\scriptsize +4.9\%} & 7.82 & 8.36 {\scriptsize +6.9\%} \\
\bottomrule
\end{tabular}
\caption{AL against the number $N$ of tokens verified per step, $T=0$, one request, fixed-500. $E$ is the number of Arbora experts, so $E=0$ is \arboraOne{} and $E=8$ is \arbora{}. The small percentages give the AL gain of $E=8$ over $E=0$, computed from unrounded values.}
\label{tab:nscan}
\end{table}

\begin{table}[ht]
\centering
\small
\begin{tabular}{@{}r r r@{}}
\toprule
Requests & DFlash2 & \arboraE{} \\
\midrule
\multicolumn{3}{@{}l}{\textit{Qwen3-8B}} \\
1 & 514.2 & \textbf{609.2} (8) \\
4 & 1{,}740.7 & \textbf{2{,}004.5} (0) \\
8 & 2{,}974.8 & \textbf{3{,}387.3} (0) \\
16 & 4{,}443.9 & \textbf{5{,}062.3} (0) \\
32 & 6{,}074.7 & \textbf{6{,}970.4} (0) \\
\midrule
\multicolumn{3}{@{}l}{\textit{Qwen3.8-27B}} \\
1 & 269.0 & \textbf{329.3} (8) \\
4 & 804.8 & \textbf{938.6} (0) \\
8 & 1{,}224.5 & \textbf{1{,}391.2} (0) \\
16 & 1{,}863.2 & \textbf{2{,}131.4} (0) \\
\bottomrule
\end{tabular}
\caption{Throughput (tok/s) on fixed-500 at $T=0$ with concurrent requests, 16 tokens verified per step, settings of Appendix~\ref{app:rounds}. The number in parentheses gives how many Arbora experts \arbora{} runs at this load. The one-request rows are the measurements of Table~\ref{tab:ablation}.}
\label{tab:concurrency}
\end{table}

\begin{table*}[t]
\centering
\footnotesize
\setlength{\tabcolsep}{2.7pt}
\begin{tabular}{@{}l r r r r r r@{}}
\toprule
& \multicolumn{3}{c}{Qwen3-8B} & \multicolumn{3}{c}{Qwen3.8-27B} \\
\cmidrule(lr){2-4}\cmidrule(l){5-7}
& DFlash2 & \shortstack{\arbora{}\\w/o\\experts} & \arbora{} & DFlash2 & \shortstack{\arbora{}\\w/o\\experts} & \arbora{} \\
\midrule
Batch preparation & 0.016 & 0.018 & 0.018 & 0.029 & 0.032 & 0.032 \\
Drafter: attention & 0.343 & 0.332 & 0.343 & 0.316 & 0.317 & 0.378 \\
Drafter: MLP & 0.617 & 0.615 & 0.822 & 0.612 & 0.611 & 0.823 \\
Drafter: output head & 0.420 & 0.420 & 0.435 & 0.428 & 0.428 & 0.010 \\
Drafter: top-16 candidates & 0.059 & 0.053 & 0.027 & 0.078 & 0.062 & 0.495 \\
Path scoring and tree & 0.017 & 0.045 & 0.076 & 0.028 & 0.048 & 0.077 \\
Other drafter kernels and gaps & 0.071 & 0.068 & 0.050 & 0.056 & 0.051 & 0.045 \\
Target verification & 6.908 & 6.836 & 6.861 & 14.785 & 14.588 & 14.549 \\
After verification & 0.154 & 0.163 & 0.167 & 0.255 & 0.253 & 0.256 \\
GPU waiting for CPU & 0.066 & 0.068 & 0.073 & 0.110 & 0.109 & 0.104 \\
\midrule
\textit{Step} & 8.673 & 8.617 & 8.873 & 16.696 & 16.498 & 16.769 \\
\bottomrule
\end{tabular}
\caption{Time per step (ms) of DFlash2, \arboraOne{}, and \arbora{}, one request, 16 tokens verified per step. Rows follow the order in which the segments run. On Qwen3.8-27B, \arbora{} computes the output head and the top-16 selection in one kernel, counted under top-16.}
\label{tab:latency}
\end{table*}

On Qwen3-8B, \arboraOne{}, \arboraEp{}, and \arbora{} are each faster with $N=64$ tokens verified per step than with $N=16$, and \arbora{} has the highest AL of the three at $N=64$ (Tables~\ref{tab:ablation} and~\ref{tab:budget}).
The main results use $N=16$, so \arbora{} verifies as many tokens per step as DFlash2.
Table~\ref{tab:nscan} gives the AL of \arboraOne{} and \arbora{} at each $N$ of Figure~\ref{fig:nscan}.
The AL gain of the Arbora experts grows with $N$ on both targets.

\subsection{High Load}
\label{app:concurrency}

Under high load, the number of Arbora experts can be reduced according to the hardware and the system load.
Table~\ref{tab:concurrency} gives the throughput of DFlash2 and \arbora{} with the number of Arbora experts that \arbora{} runs at each load.
\arbora{} is faster than DFlash2 at every number of concurrent requests.

\subsection{Per-Step Latency}
\label{app:latency}

Path scoring and tree construction take at most 0.08\,ms per step (Table~\ref{tab:latency}).
Eight Arbora experts add 0.26\,ms on Qwen3-8B and 0.27\,ms on Qwen3.8-27B to a step of \arboraOne{}, mostly in the MLP (implementation in Appendix~\ref{app:rounds}).

\section{More Workloads and Targets}
\label{app:more}

\subsection{fixed-500}
\label{app:fixed500}

Table~\ref{tab:fixed500} gives the AL and throughput of DFlash2 and \arbora{} on fixed-500, the prompts held out from the pool of the training data (\S\ref{sec:setup}).
\arbora{} has a higher AL and throughput than DFlash2 on every main target and at both $T=0$ and $T=1$.

\begin{table}[ht]
\centering
\footnotesize
\setlength{\tabcolsep}{1.7pt}
\resizebox{\columnwidth}{!}{%
\begin{tabular}{@{}l r r r r r r r r@{}}
\toprule
& \multicolumn{4}{c}{$T=0$} & \multicolumn{4}{c}{$T=1$} \\
\cmidrule(lr){2-5}\cmidrule(l){6-9}
Method & AL & $\Delta$ & tok/s & $\Delta$ & AL & $\Delta$ & tok/s & $\Delta$ \\
\midrule
\multicolumn{9}{@{}l}{\textit{Qwen3-8B}} \\
DFlash2 & 4.76 & -- & 514.2 & -- & 4.31 & -- & 464.1 & -- \\ %
\textbf{\arbora{}} & \textbf{5.98} & \textbf{+25.7\%} & \textbf{609.2} & \textbf{+18.5\%} & \textbf{5.46} & \textbf{+26.8\%} & \textbf{555.9} & \textbf{+19.8\%} \\ %
\midrule
\multicolumn{9}{@{}l}{\textit{Qwen3.8-27B}} \\
DFlash2 & 5.03 & -- & 269.0 & -- & 3.79 & -- & 206.7 & -- \\ %
\textbf{\arbora{}} & \textbf{6.40} & \textbf{+27.2\%} & \textbf{329.3} & \textbf{+22.4\%} & \textbf{4.81} & \textbf{+27.0\%} & \textbf{254.6} & \textbf{+23.2\%} \\ %
\midrule
\multicolumn{9}{@{}l}{\textit{GLM-5.3-Flash}} \\
DFlash2 & 4.43 & -- & 369.5 & -- & 3.98 & -- & 332.8 & -- \\ %
\textbf{\arbora{}} & \textbf{5.68} & \textbf{+28.2\%} & \textbf{463.8} & \textbf{+25.5\%} & \textbf{5.12} & \textbf{+28.5\%} & \textbf{418.2} & \textbf{+25.7\%} \\ %
\midrule
\multicolumn{9}{@{}l}{\textit{MiniMax-M3.1-Flash}} \\
DFlash2 & 3.59 & -- & 338.3 & -- & 3.14 & -- & 295.8 & -- \\ %
\textbf{\arbora{}} & \textbf{4.74} & \textbf{+31.9\%} & \textbf{418.3} & \textbf{+23.7\%} & \textbf{4.18} & \textbf{+33.3\%} & \textbf{370.9} & \textbf{+25.4\%} \\ %
\midrule
\multicolumn{9}{@{}l}{\textit{Mean gain over the four targets}} \\
\textbf{\arbora{}} &  & \textbf{+28.3\%} &  & \textbf{+22.5\%} &  & \textbf{+28.9\%} &  & \textbf{+23.5\%} \\ %
\bottomrule
\end{tabular}}
\caption{Results on fixed-500 with one request at a time, 16 tokens verified per step, and up to 128 generated tokens. AL includes the bonus token. $\Delta$ is the gain over DFlash2.}
\label{tab:fixed500}
\end{table}

\subsection{Domains and Public Benchmarks}
\label{app:public}

\begin{table}[ht]
\centering
\footnotesize
\setlength{\tabcolsep}{0pt}
\begin{tabular*}{\columnwidth}{@{\extracolsep{\fill}}l c c c c c@{}}
\toprule
Target & code & math & factual & chat & writing \\
\midrule
Qwen3-8B & 24.3\% & 18.2\% & 27.2\% & 26.6\% & \textbf{29.6\%} \\
Qwen3.8-27B & 27.7\% & 18.8\% & 29.1\% & \textbf{29.4\%} & 29.3\% \\
GLM-5.3-Flash & 25.0\% & 24.3\% & 28.8\% & 30.2\% & \textbf{30.9\%} \\
MiniMax-M3.1-Flash & 31.7\% & 28.4\% & 30.1\% & 32.6\% & \textbf{35.3\%} \\
\bottomrule
\end{tabular*}
\caption{AL gain of \arbora{} over DFlash2 in each domain of fixed-500 at $T=0$.}
\label{tab:domains}
\end{table}

\begin{table*}[t]
\centering
\small
\setlength{\tabcolsep}{2.6pt}
\begin{tabular}{@{}l r r r r r r@{}}
\toprule
Method & HumanEval & MT-Bench 1 & MT-Bench 2 & GSM8K & MATH-500 & MBPP \\
\midrule
\multicolumn{7}{@{}l}{\textit{Qwen3-8B, tok/s (AL)}} \\
Target alone & 155.4 & 156.2 & 150.6 & 156.0 & 156.3 & 155.8 \\
DFlash2 & 629.1 (5.48) & 485.9 (4.19) & 468.6 (4.09) & 766.9 (6.72) & 727.7 (6.35) & 630.0 (5.49) \\
\arboraE{} & \textbf{730.7} (6.66) & \textbf{590.6} (5.34) & \textbf{581.7} (5.35) & \textbf{854.2} (7.81) & \textbf{823.1} (7.51) & \textbf{729.7} (6.64) \\
\midrule
\multicolumn{7}{@{}l}{\textit{Qwen3.8-27B, tok/s (AL)}} \\
Target alone & 89.2 & 88.9 & 88.6 & 88.7 & 89.3 & 88.9 \\
DFlash2 & 299.6 (5.07) & 238.8 (4.04) & 220.0 (3.72) & 374.7 (6.63) & 351.3 (6.05) & 331.3 (5.66) \\
\arboraE{} & \textbf{371.4} (6.42) & \textbf{305.9} (5.29) & \textbf{286.1} (4.94) & \textbf{440.8} (7.98) & \textbf{416.9} (7.34) & \textbf{405.5} (7.08) \\
\midrule
\multicolumn{7}{@{}l}{\textit{GLM-5.3-Flash, tok/s (AL)}} \\
Target alone & 189.6 & 189.5 & 188.8 & 188.0 & 189.5 & 189.6 \\
DFlash2 & 562.9 (6.55) & 342.6 (3.77) & 330.7 (3.71) & 572.3 (6.80) & 615.4 (7.19) & 606.3 (7.05) \\
\arboraE{} & \textbf{676.4} (7.93) & \textbf{437.7} (4.83) & \textbf{427.3} (4.82) & \textbf{692.5} (8.43) & \textbf{732.4} (8.70) & \textbf{719.3} (8.46) \\
\midrule
\multicolumn{7}{@{}l}{\textit{MiniMax-M3.1-Flash, tok/s (AL)}} \\
Target alone & 203.5 & 203.3 & 201.5 & 200.2 & 202.9 & 203.1 \\
DFlash2 & 532.2 (5.47) & 325.8 (3.14) & 329.8 (3.27) & 504.5 (5.37) & 570.9 (5.82) & 524.0 (5.39) \\
\arboraE{} & \textbf{657.7} (7.05) & \textbf{413.0} (4.16) & \textbf{415.5} (4.30) & \textbf{612.7} (6.95) & \textbf{685.3} (7.37) & \textbf{639.7} (6.83) \\
\bottomrule
\end{tabular}
\caption{Public benchmarks at $T=0$ with one request at a time, $N=16$ tokens verified per step, up to 512 generated tokens, and thinking mode enabled. DFlash2 and \arbora{} run three rounds, reported as in Appendix~\ref{app:rounds}. MT-Bench 1 and 2 are the first and second turn, and the second turn uses the target's own first-turn answer. Target alone is the throughput of the target decoding without a drafter, measured in one round on the same GPUs, and is the denominator of the speedups in Figure~\ref{fig:overview}c. It is measured separately on each benchmark, and its throughput is nearly the same on all six.}
\label{tab:public}
\end{table*}

\begin{table*}[t]
\centering
\small
\setlength{\tabcolsep}{2.6pt}
\begin{tabular}{@{}l r r r r r r@{}}
\toprule
Method & HumanEval & MT-Bench 1 & MT-Bench 2 & GSM8K & MATH-500 & MBPP \\
\midrule
\multicolumn{7}{@{}l}{\textit{Qwen3-8B, tok/s (AL)}} \\
Target alone & 155.4 & 156.2 & 150.6 & 156.0 & 156.3 & 155.8 \\
DFlash2 & 562.9 (4.93) & 412.7 (3.57) & 398.2 (3.48) & 640.8 (5.63) & 630.3 (5.54) & 546.8 (4.78) \\
\arboraE{} & \textbf{672.4} (6.16) & \textbf{524.5} (4.70) & \textbf{500.7} (4.58) & \textbf{741.1} (6.81) & \textbf{734.3} (6.73) & \textbf{654.3} (5.99) \\
\midrule
\multicolumn{7}{@{}l}{\textit{Qwen3.8-27B, tok/s (AL)}} \\
Target alone & 89.2 & 88.9 & 88.6 & 88.7 & 89.3 & 88.9 \\
DFlash2 & 213.5 (3.56) & 175.2 (2.97) & 169.9 (2.84) & 283.1 (4.93) & 252.8 (4.33) & 225.9 (3.82) \\
\arboraE{} & \textbf{273.3} (4.66) & \textbf{223.0} (3.82) & \textbf{216.1} (3.71) & \textbf{340.3} (6.06) & \textbf{310.1} (5.41) & \textbf{284.0} (4.90) \\
\midrule
\multicolumn{7}{@{}l}{\textit{GLM-5.3-Flash, tok/s (AL)}} \\
Target alone & 189.6 & 189.5 & 188.8 & 188.0 & 189.5 & 189.6 \\
DFlash2 & 500.3 (5.78) & 301.3 (3.32) & 289.2 (3.25) & 521.9 (6.18) & 534.9 (6.24) & 521.0 (6.03) \\
\arboraE{} & \textbf{607.3} (7.12) & \textbf{393.4} (4.34) & \textbf{381.9} (4.24) & \textbf{637.2} (7.72) & \textbf{647.4} (7.63) & \textbf{632.2} (7.39) \\
\midrule
\multicolumn{7}{@{}l}{\textit{MiniMax-M3.1-Flash, tok/s (AL)}} \\
Target alone & 203.5 & 203.3 & 201.5 & 200.2 & 202.9 & 203.1 \\
DFlash2 & 428.0 (4.36) & 280.8 (2.70) & 277.8 (2.76) & 428.6 (4.52) & 484.0 (4.91) & 434.1 (4.45) \\
\arboraE{} & \textbf{544.9} (5.75) & \textbf{356.9} (3.61) & \textbf{351.8} (3.63) & \textbf{527.3} (5.88) & \textbf{588.9} (6.32) & \textbf{541.7} (5.77) \\
\bottomrule
\end{tabular}
\caption{Public benchmarks at $T=1$, in the setting of Table~\ref{tab:public}. Throughput is from the fastest of the three rounds, and AL is the mean of the three. Target alone is the $T=0$ measurement of Table~\ref{tab:public}.}
\label{tab:public-t1}
\end{table*}

\lineadj{1}{0}
\arbora{} has a higher AL and throughput than DFlash2 on HumanEval \citep{chen2021evaluating}, MT-Bench \citep{zheng2023judging}, GSM8K \citep{cobbe2021training}, MATH-500 \citep{hendrycks2021measuring,lightman2024lets}, and MBPP \citep{austin2021program}, on every main target and at both $T=0$ and $T=1$ (Tables~\ref{tab:public} and~\ref{tab:public-t1}).
At $T=0$, the AL gain of \arbora{} is largest on the two MT-Bench turns on every main target (Table~\ref{tab:public}).
On fixed-500, it is largest on creative writing or chat and smallest on math on every main target (Table~\ref{tab:domains}).

\lineadj{1}{0}
Let $\bar{p}$ be the mean probability of the target's most probable token over a prompt's output.
A low $\bar{p}$ marks a high-entropy task, because the mean entropy of the target's next-token distribution over that output is at least $-\log \bar{p}$.
On every main target, the AL gain over DFlash2 is 5.3--18.1 percentage points larger in the fifth of the fixed-500 prompts with the lowest $\bar{p}$ than in the fifth with the highest.
Math has the smallest gain of the five domains (Table~\ref{tab:domains}).
With the math prompts excluded, the difference is still 3.8--18.3 percentage points.

\subsection{More Targets}
\label{app:more-targets}

On Qwen3.6-35B-A3B, Muse-Glimmer-30B, and Gemma-4-31B-it \citep{gemmateam2026gemma4}, we run DFlash2 and \arboraOne{} with each target's released drafter (Table~\ref{tab:more-targets}).
On all three targets, \arboraOne{} has a higher AL and throughput than DFlash2.
On Qwen3.6-35B-A3B and Gemma-4-31B-it, we also train eight Arbora experts.
With them, \arbora{} raises the AL over DFlash2 by 17.9\% and 25.2\% and the throughput by 16.4\% and 18.0\%.

\begin{table}[ht]
\centering
\small
\setlength{\tabcolsep}{2.6pt}
\resizebox{\columnwidth}{!}{%
\begin{tabular}{@{}l r r r r r r@{}}
\toprule
& \multicolumn{3}{c}{AL} & \multicolumn{3}{c}{tok/s} \\
\cmidrule(lr){2-4}\cmidrule(l){5-7}
Target & DFlash2 & \shortstack{\arbora{}\\w/o\\experts} & \arbora{} & DFlash2 & \shortstack{\arbora{}\\w/o\\experts} & \arbora{} \\
\midrule
Qwen3.6-35B-A3B & 6.16 & 6.99 & \textbf{7.27} & 627.3 & 718.8 & \textbf{730.0} \\
Muse-Glimmer-30B & 7.42 & \textbf{8.42} & -- & 431.8 & \textbf{482.7} & -- \\
Gemma-4-31B-it & 5.28 & 6.28 & \textbf{6.62} & 227.5 & 262.3 & \textbf{268.4} \\
\bottomrule
\end{tabular}}
\caption{Three more targets on fixed-500 at $T=0$, with one request at a time and 16 tokens verified per step.}
\label{tab:more-targets}
\end{table}

\subsection{Released Drafters}
\label{app:released}

\lineadj{0}{1}
Table~\ref{tab:released} compares DFlash2 and \arboraOne{} on the released DFlash2 drafters of Qwen3-8B and Qwen3.8-27B, with the weights of their released path selectors and no further training.
\arboraOne{} has a higher AL and throughput than DFlash2 on both targets.
The main results use our own Qwen3-8B drafter and the further trained Qwen3.8-27B drafter, on which DFlash2 has a higher AL of 4.76 and 5.03 (Table~\ref{tab:fixed500}).
With their released weights and in their own implementations, DSpine \citep{zhang2026dspine} and XPress \citep{wang2026xpress} reach an AL of 3.16 and 3.27 on Qwen3-8B on the same prompts.

\begin{table}[ht]
\centering
\small
\setlength{\tabcolsep}{2.6pt}
\resizebox{\columnwidth}{!}{%
\begin{tabular}{@{}l r r r r@{}}
\toprule
& \multicolumn{2}{c}{AL} & \multicolumn{2}{c}{tok/s} \\
\cmidrule(lr){2-3}\cmidrule(l){4-5}
Target & DFlash2 & \arboraOne{} & DFlash2 & \arboraOne{} \\
\midrule
Qwen3-8B & 2.74 & \textbf{3.44} & 308.0 & \textbf{378.2} \\
Qwen3.8-27B & 4.84 & \textbf{5.81} & 262.3 & \textbf{311.9} \\
\bottomrule
\end{tabular}}
\caption{Released drafters on fixed-500, with one request at a time.}
\label{tab:released}
\end{table}

\subsection{GRAFT and TAPS}
\label{app:tree-methods}

Table~\ref{tab:tree-methods} compares \arbora{} with GRAFT \citep{ye2026graft} and TAPS \citep{wang2026taps}, two tree methods that also score a token given the token before it (\S\ref{sec:related}).
\arbora{} has the highest AL in every column, ahead of TAPS, GRAFT, and DDTree \citep{ringel2026ddtree}.

GRAFT and TAPS run on the same shared drafter as \arbora{}, verify the same $N=16$ tokens per step, and use as many candidates per position as one draft of \arbora{} (32 on MiniMax-M3.1-Flash, 16 on the other targets).
Each is trained on the target's responses to 20{,}000 training prompts, as the path scorers of Table~\ref{tab:scorer-lr} are (Appendix~\ref{app:treeloss}).
The learning rate of each method is chosen on held-out requests among the four values of Appendix~\ref{app:treeloss} and the value from its paper or code.
On Qwen3-8B and Qwen3.8-27B, the AL of GRAFT and TAPS is trace-driven.
All rows of Table~\ref{tab:tree-methods} use the drafter, the path scorer, and the Arbora experts with which the traces were recorded, so every method is compared on the same weights.
These weights are not those of Table~\ref{tab:fixed500}, so the AL of \arbora{} differs between the two tables.
For these weights, the trace-driven AL is within 0.03 of the serving AL on Qwen3.8-27B and about 2\% higher on Qwen3-8B (Appendix~\ref{app:offline}).
Both differences are smaller than the margin of \arbora{} over GRAFT and TAPS on the same target.

\begin{table}[ht]
\centering
\small
\setlength{\tabcolsep}{4pt}
\begin{tabular}{@{}l r r r r@{}}
\toprule
Method & \shortstack{Qwen3-\\8B} & \shortstack{Qwen3.8-\\27B} & \shortstack{GLM-5.3-\\Flash} & \shortstack{MiniMax-\\M3.1-Flash} \\
\midrule
DDTree & 5.078 & 5.517 & 4.853 & 4.096 \\
GRAFT & 5.445 & 5.808 & 4.881 & 4.217 \\
TAPS & 5.522 & 5.825 & 5.231 & 4.393 \\
\textbf{\arbora{}} & \textbf{5.720} & \textbf{6.305} & \textbf{5.608} & \textbf{4.750} \\
\bottomrule
\end{tabular}
\caption{AL at $T=0$ on fixed-500 with 16 tokens verified per step. DDTree, GRAFT, and TAPS run on each target's shared drafter. GRAFT and TAPS are trace-driven on Qwen3-8B and Qwen3.8-27B. All other entries are measured in serving with one request at a time.}
\label{tab:tree-methods}
\end{table}

\section{Learning a New Domain Online}
\label{app:online}

\paragraph{Setting.}
The stream is 20{,}000 software-\hspace{0pt}engineering (SWE) requests.
They are SWE-agent turns from SWE-smith \citep{yang2025swesmith}, shuffled once and served in order.
Qwen3-8B answers them at $T=0$, one request at a time, with thinking mode disabled, contexts cut to 2{,}560 tokens, and up to 512 new tokens.

\begin{figure}[ht]
\centering
\includegraphics[width=\columnwidth]{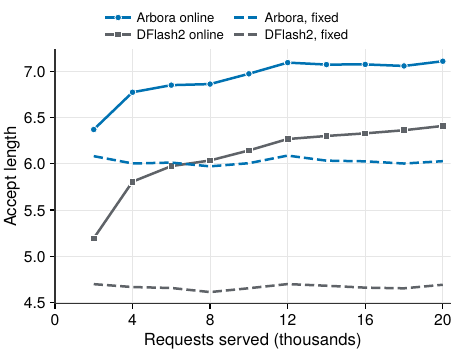}
\caption{AL on the SWE stream per 2{,}000 requests, on Qwen3-8B with one request at a time. \arbora{} online adds a new Arbora expert and trains it and the path scorer. DFlash2 online trains its drafter at learning rate $10^{-5}$, as in \S\ref{sec:online}.}
\label{fig:online}
\end{figure}

\begin{table}[ht]
\centering
\small
\setlength{\tabcolsep}{1.6pt}
\begin{tabular}{@{}l r r r r r@{}}
\toprule
Requests (thousands) & 0--4 & 4--8 & 8--12 & 12--16 & 16--20 \\
\midrule
DFlash2, fixed & 4.68 & 4.63 & 4.68 & 4.67 & 4.67 \\
DFlash2 online, $10^{-5}$ & 5.48 & 6.00 & 6.21 & 6.31 & 6.39 \\
\quad $3\times10^{-5}$ & 5.62 & 6.07 & 6.22 & 6.30 & 6.26 \\
\quad $10^{-4}$ & 5.76 & 6.33 & 6.50 & 6.42 & 6.51 \\
\arbora{}, fixed & 6.04 & 5.99 & 6.05 & 6.03 & 6.02 \\
\arbora{} online: path scorer & 6.46 & 6.66 & 6.79 & 6.84 & 6.84 \\
\quad new Arbora expert & 6.42 & 6.57 & 6.71 & 6.72 & 6.74 \\
\quad both & 6.57 & 6.86 & 7.03 & 7.07 & \textbf{7.08} \\
\bottomrule
\end{tabular}
\caption{AL on the SWE stream. Each row is a separate run. ``fixed'' is not trained online. \arbora{} online trains its path scorer, one new Arbora expert, or both. The learning rates shown are those of DFlash2's drafter.}
\label{tab:online-stream}
\end{table}

\begin{table*}[ht]
\centering
\small
\setlength{\tabcolsep}{3pt}
\begin{tabular*}{\textwidth}{@{\extracolsep{\fill}}l r r r r r r r r r@{}}
\toprule
& \multicolumn{2}{c}{Start} & \multicolumn{2}{c}{After the stream} & \multicolumn{5}{c@{}}{Other domains, change} \\
\cmidrule(lr){2-3}\cmidrule(lr){4-5}\cmidrule(l){6-10}
Trained online & SWE & other & SWE & other & code & writing & factual & chat & math \\
\midrule
DFlash2: drafter, $10^{-5}$ & 4.50 & 4.76 & 5.99 & 4.55 ($-$4.4\%) & $-$6.1\% & $-$4.6\% & $-$3.0\% & $-$4.9\% & $-$3.6\% \\
\quad drafter, $3\times10^{-5}$ & 4.50 & 4.76 & 5.89 & 4.52 ($-$5.1\%) & $-$6.2\% & $-$4.8\% & $-$3.9\% & $-$5.5\% & $-$5.4\% \\
\quad drafter, $10^{-4}$ & 4.50 & 4.76 & 6.02 & 4.43 ($-$7.0\%) & $-$9.6\% & $-$5.9\% & $-$5.3\% & $-$7.4\% & $-$7.3\% \\
\midrule
\arbora{}: path scorer & 5.78 & 5.98 & 6.54 & 5.87 ($-$1.8\%) & $-$1.7\% & $-$1.4\% & $-$3.2\% & $-$1.2\% & $-$1.3\% \\
\arbora{}: new Arbora expert & 5.83 & 6.02 & 6.54 & 6.02 (+0.1\%) & 0.0\% & +0.9\% & 0.0\% & +0.4\% & $-$1.1\% \\
\arbora{}: both & 5.82 & 6.02 & \textbf{6.84} & 5.92 ($-$1.6\%) & $-$2.1\% & $-$0.7\% & $-$1.0\% & $-$1.8\% & $-$2.8\% \\
\bottomrule
\end{tabular*}
\caption{AL on held-out data before and after the 20{,}000-request stream, on Qwen3-8B with one request at a time. Each row is a separate online run. SWE is 500 SWE-bench Verified turns, and other is the five domains of fixed-500 together. Percentages are relative to the row's own start. The start differs between rows that serve different numbers of Arbora experts. ``both'' trains the path scorer and the new Arbora expert. The DFlash2 rows train the drafter at the learning rate shown and the path selector at $10^{-4}$.}
\label{tab:online-ablation}
\end{table*}

Each online run covers the whole stream.
It serves on one H100 and trains on two others.
After every request, the server sends the prompt and answer tokens to the trainer.
The trainer recomputes the target's hidden states on its own GPUs and trains asynchronously.
Every 120\,s, the server loads the new weights in place, without a restart.
An \arbora{} update is about 346\,MB and takes 0.107\,s on average to load, against 2.50\,GB and 0.628\,s for DFlash2.

\arbora{} keeps its shared drafter and its eight Arbora experts frozen and trains two parts online.
The first is one new Arbora expert of the same width as the others, randomly initialized and trained with the loss of Eq.~\ref{eq:mcl}.
With it, \arbora{} serves nine Arbora experts.
The second is the path scorer, trained with the loss of Eq.~\ref{eq:treeloss} on the shared drafter's candidates.
The learning rates are $10^{-3}$ for the new Arbora expert and $10^{-4}$ for the path scorer, the best in a simulation of the stream.
They equal the learning rates of the offline training of the Arbora experts on Qwen3-8B and of the path scorer (Appendices~\ref{app:training} and~\ref{app:treeloss}).

DFlash2 trains its whole drafter and its path selector online, as Aurora does.
On the stream, we run DFlash2 at drafter learning rates $10^{-5}$, $3\times10^{-5}$, and $10^{-4}$, with the path selector at $10^{-4}$.
The three reach a similar AL on SWE-bench Verified.
\S\ref{sec:online} reports $10^{-5}$, which loses the least on the other domains.

Two further \arbora{} runs train one part each.
One trains only the path scorer and serves the eight Arbora experts without a new one.
The other trains only the new Arbora expert and serves all nine.

\paragraph{Results.}
Figure~\ref{fig:online} and Table~\ref{tab:online-stream} give the AL on the stream.
Table~\ref{tab:online-ablation} gives the AL before and after the stream on held-out data, which are 500 SWE-bench Verified turns \citep{jimenez2024swebench,chowdhury2024verified} from repositories outside the stream and the five domains of fixed-500.
Training only the new Arbora expert keeps each other domain within 1.1\% of its start.
The small loss on the other domains when the path scorer is trained as well comes from the update of the path scorer.
Each part alone raises the AL on SWE-bench Verified to 6.54, and both together raise it to 6.84.
DFlash2, which updates its whole drafter, loses more on the five other domains together than every \arbora{} run.
At each of its three learning rates, DFlash2 also has a lower AL than both \arbora{} settings of \S\ref{sec:online} in every column of Table~\ref{tab:online-stream} and on SWE-bench Verified.

\section{Artifacts, Licenses, and Compute}
\label{app:artifacts}

\paragraph{Licenses.}
Qwen3-8B, Qwen3.8-27B, Qwen3.6-35B-A3B, Muse-Glimmer-30B, and Gemma-4-31B-it are licensed under Apache-2.0.
GLM-5.3-Flash is licensed under MIT, and MiniMax-M3.1-Flash under the MiniMax Community License.
The Qwen3-8B DFlash drafter is licensed under MIT.
The DFlash2 drafters for Qwen3-8B, Qwen3.8-27B, Qwen3.6-35B-A3B, Muse-Glimmer-30B, Gemma-4-31B-it, and MiniMax-M3.1-Flash are licensed under Apache-2.0.
The GLM-5.3-Flash DFlash2 drafter is licensed under CC BY-NC-ND 4.0.
We use it for non-commercial research only and do not distribute weights derived from it.
These weights include the path scorer trained from its path selector and the Arbora experts trained on it.
HumanEval and GSM8K are licensed under MIT, and MBPP under CC-BY-4.0.
The 80 MT-Bench prompts (HuggingFaceH4/mt\_bench\_prompts, taken from FastChat) are licensed under Apache-2.0.
MATH-500 is taken from PRM800K and MATH, which are both licensed under MIT.
All models and data are used for research on inference speed, consistent with their licenses.

\paragraph{Parameters.}
The counts exclude embeddings and output heads, which are frozen.
The Qwen3-8B and GLM-5.3-Flash drafters each have 1.05B parameters, and DFlash2's convolutions add 42M.
Their path scorers have 79M and 80M, and their eight Arbora experts add 755M.
The path scorer has the same parameters as DFlash2's path selector, so only the Arbora experts add parameters.
They are stored in INT4 and add 0.26\,ms to a step on Qwen3-8B (Appendices~\ref{app:rounds} and~\ref{app:latency}).
The Qwen3.8-27B drafter has 1.73B parameters, and the convolutions add 66M.
Its path scorer has 128M, and its eight Arbora experts add 1.34B.
The MiniMax-M3.1-Flash drafter has 1.93B parameters, and the convolutions add 95M.
Its path scorer has 104M, and its eight Arbora experts add 1.13B.

\paragraph{Compute.}
All training ran on H100 GPUs.
The target ran on separate GPUs and supplied its hidden states.
On Qwen3-8B, training used eight GPUs, two of them for the target.
The DFlash2 drafter took 16.1 hours, the path scorer 4.9 hours, and the Arbora experts 7.7 hours.
On Qwen3.8-27B, training used eight GPUs, four of them for the target.
The path scorer took 11.2 hours and the Arbora experts 8.0 hours.
The three additional epochs of the Qwen3.8-27B drafter took about 65 hours on eight GPUs.
On GLM-5.3-Flash, training used six GPUs, and eight more ran the target.
The path scorer took 3.1 hours and the Arbora experts 4.2 hours.
On MiniMax-M3.1-Flash, training used eight GPUs, four of them for the target.
The path scorer took 4.0 hours and the Arbora experts 8.7 hours.
On Qwen3.6-35B-A3B and Gemma-4-31B-it, training used eight GPUs, four of them for the target.
The path scorers took 1.7 and 1.9 hours and the Arbora experts 4.9 and 9.7 hours.
These times exclude generating the training data and the serving measurements.

\end{document}